\documentclass{article} %
\usepackage[preprint]{colm2026_conference}

\usepackage{amsmath,amsfonts,bm}

\def\eqref#1{equation~\ref{#1}}

\def\1{\bm{1}}

\DeclareMathAlphabet{\mathsfit}{\encodingdefault}{\sfdefault}{m}{sl}
\SetMathAlphabet{\mathsfit}{bold}{\encodingdefault}{\sfdefault}{bx}{n}

\usepackage{microtype}
\usepackage{xspace}
\usepackage{graphicx}
\usepackage{booktabs}
\usepackage{hyperref}
\hypersetup{
    colorlinks,
    linkcolor={red!50!black},
    citecolor={blue!50!black},
    urlcolor={blue!80!black}
}
\usepackage{url}
\usepackage{makecell}
\usepackage{multirow}
\usepackage{adjustbox}
\usepackage[table]{xcolor}
\usepackage{enumitem}
\usepackage{algorithm}
\usepackage{algpseudocode}
\usepackage[most]{tcolorbox}
\usepackage[T1]{fontenc}
\usepackage{booktabs}
\usepackage[table]{xcolor}
\usepackage{array}
\usepackage{tabularx}
\usepackage{pifont} %
\usepackage{caption}

\usepackage{lineno}

\definecolor{darkblue}{rgb}{0, 0, 0.5}
\hypersetup{colorlinks=true, citecolor=darkblue, linkcolor=darkblue, urlcolor=darkblue}

\newcommand{\xmark}{\ding{53}} %
\newcommand{\cmark}{\ding{109}} %
\newcommand{\hquad}{\hspace{0.5em}} 

\newtcolorbox[auto counter, 
  number freestyle={\noexpand\arabic{\tcbcounter}}]{mybox}[2][]{%
    enhanced,
    breakable,
    fonttitle=\bfseries,
    title=Box~\thetcbcounter: #2,
    #1
}

\newcommand{\privasis}{\textsc{Privasis}\xspace}
\newcommand{\sanitizationdataset}{\textsc{Privasis-Sanitization}\xspace}
\newcommand{\sanitizer}{\textsc{Privasis-Cleaner}\xspace}

\title{Privasis: Synthesizing the Largest\\``Public'' Private Dataset from Scratch}

\author{Hyunwoo Kim$^{1*}$ \hquad Niloofar Mireshghallah$^{2*}$ \hquad Michael Duan$^{3}$ \hquad Rui Xin$^{4}$ \\
\textbf{Shuyue Stella Li$^{4}$ \quad Jaehun Jung$^{1}$ \quad David Acuna$^{1}$ \quad Qi Pang$^{1}$ \quad Hanshen Xiao$^{1}$}\\
\textbf{G. Edward Suh$^{1}$ \quad Sewoong Oh$^{4}$ \quad Yulia Tsvetkov$^{4}$ \quad Pang Wei Koh$^{4}$ \quad Yejin Choi$^{1}$}\\
\\
$^{1}$NVIDIA \qquad $^{2}$CMU \qquad $^{3}$USC \qquad $^{4}$UW
}

\definecolor{forestgreen}{rgb}{0.13,0.55,0.13}

\begin{document}

\ifcolmsubmission
\linenumbers
\fi

\maketitle

\begin{abstract}
Research involving privacy-sensitive data has always been constrained by data scarcity, standing in sharp contrast to other areas that have benefited from data scaling.
This challenge is becoming increasingly urgent as modern AI agents—such as OpenClaw and Gemini Agent—are granted persistent access to highly sensitive personal information.
To tackle this longstanding bottleneck and the rising risks, we present \privasis (\textit{i.e., privacy oasis}), the first million-scale fully synthetic dataset entirely built from scratch---an expansive reservoir of texts with rich and diverse private information---designed to broaden and accelerate research in areas where processing sensitive social data is inevitable.
Compared to existing datasets, \privasis, comprising 1.4 million records, offers orders-of-magnitude larger scale with quality, and far greater diversity across various document types, including medical history, legal documents, financial records, calendars, and text messages with a total of 55.1 million annotated attributes such as ethnicity, date of birth, workplace, etc.
We leverage \privasis to construct a parallel corpus for text sanitization with our pipeline that decomposes texts and applies targeted sanitization.
Our compact sanitization models ($\leq$4B) trained on this dataset outperform state-of-the-art large language models, such as GPT-5 and Qwen-3 235B.
We plan to release data, models, and code to accelerate future research on privacy-sensitive domains and agents.\footnote{\url{https://privasis.github.io}}
\end{abstract}

\section{Introduction}

\begin{figure}[ht]
    \centering
    \includegraphics[width=\linewidth]{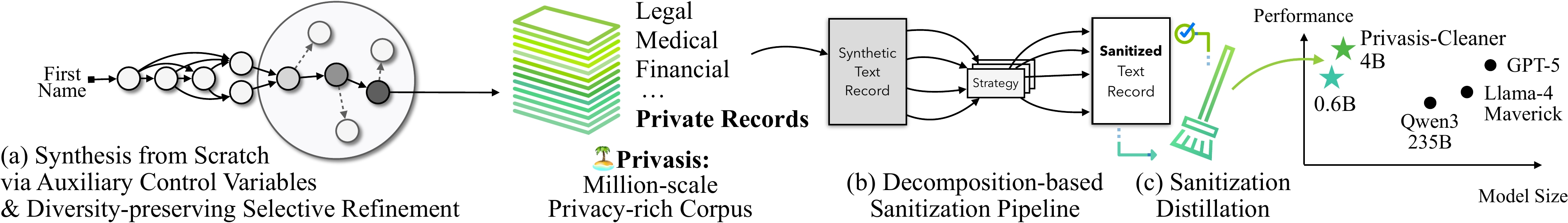}
    \vspace{-5pt}
    \caption{\textbf{\privasis, the Privacy Oasis Dataset}:
    We synthesize the first publicly-available million-scale dataset with diverse private information, entirely from scratch.
    (a) Using auxiliary control variables, we initialize a text record draft containing rich private information and then selectively refine it while preserving the overall diversity of the record set (\S \ref{sec:synthesis}).
    (b) Based on this, we construct a parallel corpus for text sanitization using a decomposition-based sanitization pipeline (\S \ref{sec:sanitization}).
    (c) On this parallel corpus, we train compact sanitization models ($\leq$4B) that outperform GPT-5 (\S \ref{sec:experiments}).
    } 
    \label{fig:overview}
\end{figure}

Progress in privacy-related research has long been fundamentally limited by a drought of data. By definition, private information cannot be publicly shared. 
As a result, most prior work relies on small, narrowly scoped datasets—standing in stark contrast to the data-driven scaling paradigm that underpins progress in many other areas of AI~\citep{li-etal-2025-papillon,yukhymenko2024synthpai}.
Meanwhile, agentic systems (e.g., \citealp[OpenClaw;][]{openclaw}, \citealp[Gemini Agent;][]{gemini_agent}, and \citealp[ChatGPT Health;][]{chatgpt_health}) increasingly need to process personal communications, documents, and records at inference time, while maintaining privacy guarantees \citep{mireshghallah2024confaide}.
This trend highlights the urgent need for robust privacy methods at multiple stages: input-side approaches like data sanitization and minimization \citep{zhou2025rescriber,dou-etal-2024-reducing}, as well as post-hoc techniques \citep{bagdasarian2024airgapagent} that ensure models appropriately handle the personal information entrusted to them.
Yet despite their apparent simplicity, these privacy tasks remain surprisingly difficult—current LLMs fail at even basic personally identifiable information (PII) detection \citep{shao2024privacylens,pham2025can}.

To address this urgent need, we introduce \privasis (\textit{i.e., privacy oasis}), the first million-scale synthetic dataset built entirely from scratch for privacy research along with its corresponding parallel corpus \sanitizationdataset for text sanitization (Figure~\ref{fig:overview}). Our synthesis pipeline (Figure~\ref{fig:overview}a; \S \ref{sec:synthesis}) achieves this scale without reference data by using auxiliary control variables—profiles with personal attributes (e.g., name, ID), record types (e.g., ``psychotherapy billing statement''), and background contexts—to generate diverse documents spanning medical, legal, financial, and communication records, each annotated with detailed JSON structures. To ensure realism and diversity, we employ iterative rejection sampling with a weighted criterion combining LLM quality scoring and Vendi diversity metrics~\citep{friedman2023vendi}. \privasis demonstrates superior diversity compared to existing human-written datasets: our domain subsets consistently achieve higher MATTR (lexical diversity metric; 0.807--0.823~vs.~0.700--0.794), bigram diversity, and Shannon entropy while maintaining lower cosine similarity, indicating richer lexical variety and reduced semantic redundancy.

We evaluate models on a new benchmark derived from \privasis by testing their ability to detect and sanitize private information across both vanilla and harder test sets. 
Even frontier models leave room for improvement: GPT-5 achieves only 70\% and 13\% full success rate on vanilla and hard sets, respectively. 
To tackle this challenge, we design \sanitizationdataset, a parallel corpus for training models to selectively remove or abstract sensitive information while preserving textual utility (\S \ref{sec:sanitization}).
Our decomposition-based pipeline (Figure~\ref{fig:overview}b) breaks records into chunks, then applies targeted sanitization based on user-specified attributes—going beyond fixed PII categories to support arbitrary information that users may contextually want removed.

Our decomposition-based pipeline supports multiple abstraction levels (e.g., replacing ``March 3rd'' with ``Early March''~vs.~complete removal) and explicitly preserves non-sensitive information through retention targets, yielding triplets of (original record, sanitization instruction, sanitized record) that enable training lightweight models for flexible, utility-preserving sanitization. Training on \sanitizationdataset yields compact models that outperform frontier LLMs (\S \ref{subsec:exp_results}): our 4B-parameter \sanitizer achieves 72.5\% full success rate on the vanilla test set, surpassing all tested models including o3 (70.3\%), while maintaining competitive performance on the hard set (12.4\%~vs.~GPT-5's 13.1\%). Crucially, these compact models enable practical on-device data minimization—removing unnecessary sensitive information before processing—which is essential since users cannot risk sending private data to external servers for cleaning~\citep{zhou2025rescriber}.

\privasis provides the first privacy-safe yet privacy-rich dataset at million scale, overcoming the fundamental data scarcity bottleneck in privacy research. Unlike prior work that relies on real-world reference data or repurposed existing datasets, \privasis is entirely reference-free—synthesized from scratch using only auxiliary control variables and public name databases—eliminating privacy risks from actual individuals. We validate this privacy safety by sampling more than 1K profiles and querying whether they correspond to real people: while some shared names or partial attributes with real individuals, manual verification revealed no genuine matches, with generated profiles being hallucinated rather than memorized from training data. 
With its rich records and attributes, future work can leverage \privasis to develop methods that respect privacy by design—from improved sanitization models to differential privacy techniques, and agentic systems that must operate responsibly on sensitive information.
We plan to release all code, data, and models to accelerate progress in this critical area where technical capability must align with ethical responsibility.

\section{Synthesizing Privacy-rich Text Data from Scratch}
\label{sec:synthesis}

\begin{figure}[t!]
    \centering
    \includegraphics[width=\linewidth]{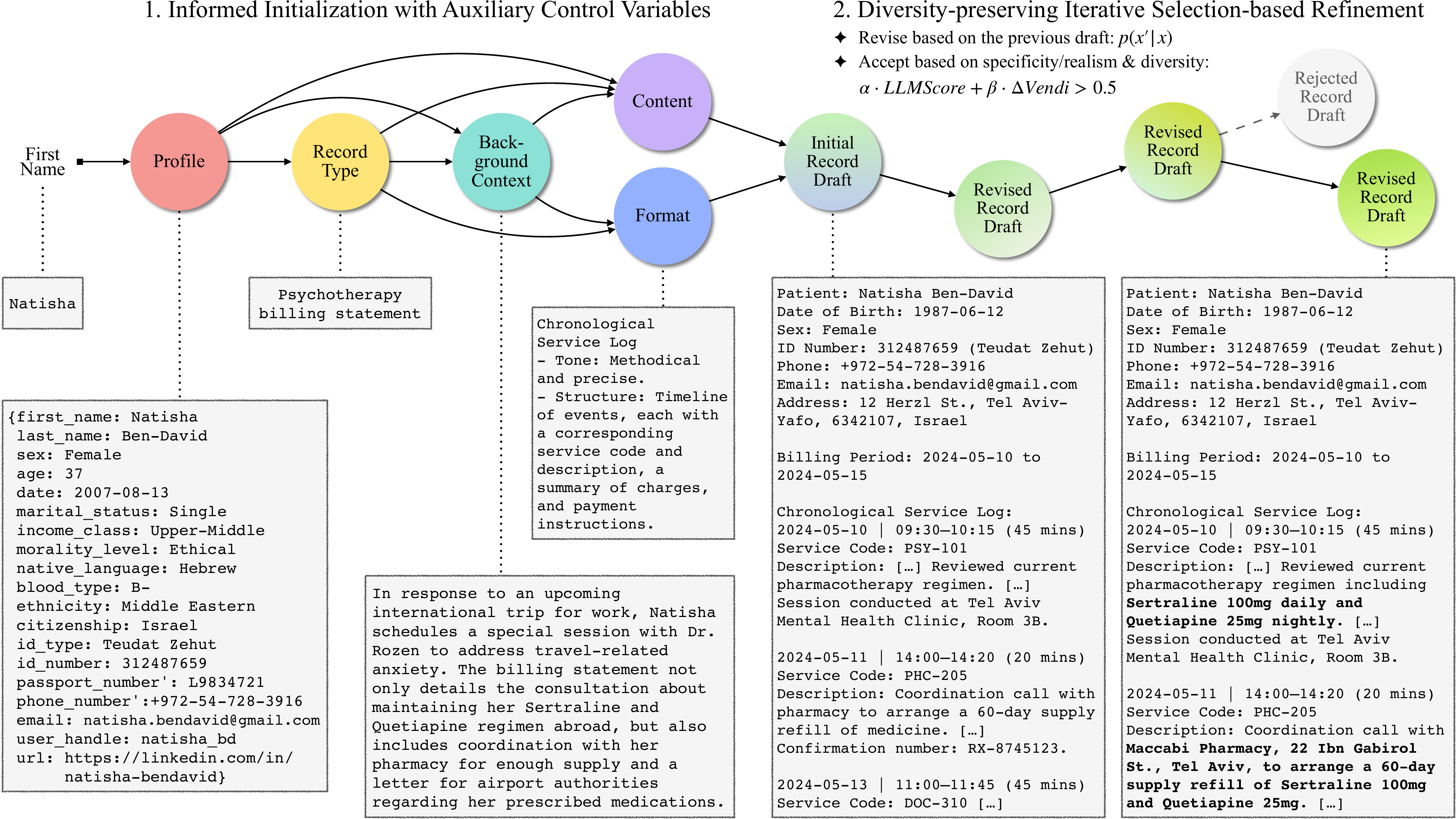}
    \vspace{-15pt}
    \caption{
    Overview of our synthesis pipeline.
    } 
    \vspace{-5pt}
    \label{fig:synthesis_pipeline}
\end{figure}

The construction of \privasis is guided by three design principles:  
(1) scalable synthesis across a broad spectrum of text records,  
(2) incorporation of diverse and fine-grained private information within those records, and 
(3) synthesis that does not rely on real-world reference data. 

To this end, we use LLMs as they define expressive probability distributions over text.
However, directly sampling such complex, specific data $x$ from LLMs is challenging, as they tend to favor high-probability, generic continuations rather than rare, highly specific instances of private information.
This is particularly difficult without reference data, which most existing works rely on to steer generations.
To address this challenge, we adopt informed initialization through auxiliary control variables, followed by a diversity-preserving revision algorithm with selection.  
This allows us to efficiently explore the large space of possible texts even in the absence of reference data.
Figure \ref{fig:synthesis_pipeline} provides an overview of our pipeline.  
More details and examples are in Appendix \ref{app:synthesis} and \ref{app:examples}.

\subsection{Synthesis Pipeline}
\label{subsec:synthesis_pipeline}

\noindent\textbf{1. Informed Initialization with Auxiliary Control Variables:}
We introduce multiple auxiliary control variables to guide the initialization of a text record $x$.
A record is determined by two primary variables—\textit{semantic content} ($c$) and \textit{structural format} ($f$)—which are themselves informed by three auxiliary variables (Figure \ref{fig:synthesis_pipeline}):
\begin{itemize}[leftmargin=*, nosep]
\item Profile ($i$): Basic attributes such as gender, ethnicity, and date, sampled from a predefined set conditioned on the first name sampled from the US SSN applicant database.\footnote{\href{https://catalog.data.gov/dataset/baby-names-from-social-security-card-applications-national-data} 
{\scriptsize{\nolinkurl{catalog.data.gov/dataset/baby-names-from-social-security-card-applications-national-data}}}} The profile also includes attributes describing a specific event involving the individual.
\item Record type ($d$): Concise description of what the record is, derived from $i$.
\item Background context ($b$): Description of the social context of the record, derived from $i$ and $d$.
\end{itemize}
To encourage diversity, we prompt the LLM to generate multiple candidates for $d$ and $b$ in a list format and then select one at random.
The record's semantic content ($c$) is constructed as the concatenation of $i$, $d$, and $b$, while its format ($f$) is generated given $d$ and $b$.
Finally, the initial draft $x_0$ is generated, given $c$ and $f$.
Because the process is bottom-up from explicit auxiliary variables, the variables serve as free annotations or metadata alongside the record.

\noindent\textbf{2. Diversity-preserving Iterative Selection-based Refinement:}
The initial draft $x_0$ may contain degenerate or overly generic content, while our goal is to produce records with realistic and concrete details.
To improve the quality of $x_0$, we iteratively apply selective refinement.
At each step $t$, a candidate draft $x'_t$ is sampled and evaluated against the current draft $x_t$.
An LLM judge compares which of the two drafts are better in terms of specificity and realism.

We find repeated refinement leads to a lack of variety converging to similar patterns.
To mitigate this, we use the Vendi embedding score \citep{friedman2023vendi}, which measures how spread out a set of representations is in embedding space.
Intuitively, the score is higher when records cover a broader range of semantic directions, and lower when they collapse to similar content.
Concretely, we maintain a collection of all final accepted records produced so far, and for each diversity evaluation we randomly sample up to $n_P$ records from this collection to form a pool $P$.
The contribution of a new draft to diversity is defined as the change in Vendi score between $P \cup \{x_t\}$ (with the current draft) and $P \cup \{x_t'\}$ (with the new draft).
The final decision is based on a weighted acceptance score, 
\begin{equation*}
S(x'_t) = \alpha \cdot \text{LLMScore}(x_t, x'_t) + \beta \cdot \bigl(\text{Vendi}(P \cup x_t') - \text{Vendi}(P \cup x_t)\bigr)
\end{equation*}
where the candidate is accepted only if $S(x'_t) > \tau$ with $\tau = 0.5$.
The refinement procedure is repeated for up to three steps.
Ablation study of our pipeline are provided in Appendix \ref{app:subsec:privasis_analysis}.

\noindent\textbf{3. Attribute Annotation:}  
After the final refinement step, we extract and annotate additional attributes in JSON format that are present in the record but not explicitly captured in the profile $i$ (e.g., fine-grained details mentioned in the text).
These attributes and $i$ are then grouped into semantic clusters using an LLM (i.e., \textit{grouped attributes}).  
For example in medical records, `clinic name', `pharmacy name', and `room number' clustered under `location'.
Such groupings yield contextual structure that can be leveraged in downstream tasks, including sanitization (\S \ref{subsec:sanitization_pipeline}).  

\noindent\textbf{4. Filtering:}  
We filter out cases where any type of error occurred during the generation process, yielding a success rate of approximately 94\%.
Because the initial drafts of the records are refined, most appear reasonable.
Nonetheless, we exclude records with fewer than 64 words (66,894 in total), profiles with an age under 18 (1,180), and degenerate cases (1,232).

\subsection{Statistics \& Analysis}
\label{subsec:privasis_analysis}

\noindent\textbf{Basic Statistics:}
\privasis comprises 1,414,871 records with 55,092,084 annotated attributes ($\approx$39 per record). These span basic profile details (e.g., name, sex, age, marital status, income, language, blood type, phone, email, URL) as well as richer information such as dates and locations. Attributes are grouped into an average of 6.2 clusters per record, each with 6.3 attributes. Records also include background context, format, and type descriptions averaging 527.0, 76.4, 41.8, and 20.0 words, respectively.
Most records are generated by GPT-OSS-120B \citep[67.9\%;][]{openai2025gptoss}, followed by GPT-4.1-Mini (21.6\%) and Exaone-3.5-32B \citep[7.2\%;][]{research2024exaone}. Smaller shares come from Qwen3-80B \citep[1.1\%;][]{yang2025qwen3}, Llama-3.3-70B \citep[1.1\%;][]{grattafiori2024llama}, GPT-4.1 (0.7\%), and other frontier models (<0.1\%). Using multiple models both increases stylistic and distributional diversity and shows that our pipeline generalizes across LLMs.

\begin{table}[t!]
\caption{Distribution of domain categories in \privasis with the top three subcategories for each main category. Percentages represent the proportion of each category among sampled records.}
\centering
\scriptsize
\renewcommand{\arraystretch}{1.05}
\begin{adjustbox}{width=1.0\linewidth}
\begin{tabular}{lr|lll}
\toprule
\textbf{Domain Category} & \textbf{Ratio} & \textbf{Subcategory 1} & \textbf{Subcategory 2} & \textbf{Subcategory 3} \\
\midrule
Health \& Wellness & 20.7\% & Medical Care (11.8\%) & Mental Health \& Support (4.2\%) & Healthcare Administration (3.2\%) \\
Government \& Civic & 13.5\% & Immigration \& Citizenship (4.9\%) & Legal Proceedings (3.8\%) & Public Administration (3.2\%) \\
Business \& Finance & 13.4\% & Employment \& HR (4.9\%) & Financial Management (4.3\%) & Accounting \& Tax (1.9\%) \\
Personal \& Family & 10.7\% & Personal Development (6.0\%) & Family Relationships (2.0\%) & Life Events (1.9\%) \\
Community \& Social & 9.3\% & Community Services (4.8\%) & Religious \& Spiritual (2.3\%) & Volunteer \& Nonprofit (1.1\%) \\
Professional Services & 9.1\% & Project Management (6.0\%) & Consulting \& Advisory (1.1\%) & Specialized Services (1.0\%) \\
Education \& Training & 8.9\% & Academic Administration (5.6\%) & Financial Aid (1.8\%) & Learning \& Development (0.8\%) \\
Legal \& Compliance & 7.6\% & Criminal Justice (2.3\%) & Contracts \& Agreements (2.1\%) & Court \& Litigation (1.6\%) \\
Media \& Comms. & 3.9\% & Publishing \& Content (1.6\%) & Creative Arts (1.3\%) & Communication (0.7\%) \\
Recreation \& Lifestyle & 2.3\% & Travel \& Tourism (1.2\%) & Food \& Culinary (0.5\%) & Entertainment \& Hobbies (0.5\%) \\
Technical \& Operations & 0.7\% & System Administration (0.6\%) & Technical Support (0.1\%) & -- \\

\bottomrule
\end{tabular}
\end{adjustbox}
\label{tab:domain_distribution}
\end{table}

\noindent\textbf{Distribution of Generated Profiles:}
The sex distribution is 59\% female and 41\% male. 
The age, income class, and blood type distributions are uniform, as these attributes were randomly sampled.
The ethnicity distribution is shown in Figure \ref{fig:ethnicity} in the Appendix.
The most common ethnicity is South Asian, followed by European and Black.
It should be noted that these categories are generated by the LLMs; hence, the granularity of the ethnicity labels is heterogeneous.
For example, European refers to a region-based category, whereas Black is defined in terms of phenotype.
We also found a large proportion of profiles were non-U.S. (95\%).
This is because we only use first names from the U.S. SSN database, which reflects a highly diverse population.

\noindent\textbf{Category Distribution of Generated Records:}
Although records include annotated \textit{record types}, these are often overly specific (e.g., \textit{psychotherapy billing statement from Dr. Rozen}), making clustering difficult. We therefore re-categorize them into broader groups (e.g., \textit{medical care}) using GPT-4.1-Mini on a random sample of 48K records (prompt in Appendix~\ref{app:subsec:privasis_analysis}), yielding 592 unique categories. These were hierarchically clustered with Claude Sonnet 4 and manually refined into 10 primary categories and 42 subcategories.
Table \ref{tab:domain_distribution} shows the distribution of the main categories along with the top 3 subcategories within each of the main category.
\textit{Health \& Wellness} is the most common category (20.7\%), followed by \textit{Government \& Civic} (13.5\%) and \textit{Business \& Finance} (13.4\%).
See Appendix~\ref{app:examples} for example records and metadata (e.g., attributes) per category.

\noindent\textbf{How does \privasis compare to human-written datasets?}
Table~\ref{tab:diversity} reports four quantitative diversity metrics: Moving-average TTR \citep[MATTR;][]{covington2010mattr}, bigram diversity, Shannon entropy, and cosine similarity.
We compare each \privasis domain subset with its related human-written dataset.
\privasis subsets consistently exhibit greater diversity than human-written datasets across multiple metrics.
MATTR and bigram diversity are higher across \privasis domains, reflecting richer vocabulary and syntactic variation, while higher Shannon entropy indicates more uniform word use and less repetition.
Additionally, \privasis subsets achieve lower cosine similarity scores, confirming decreased redundancy and greater semantic diversity within the dataset.

We also conducted a human evaluation of the naturalness and coherence of the records in \privasis.
Specifically, we randomly sampled 128 records from \privasis and 128 records from the collection of human-written datasets spanning a wide range of domains similar to those in \privasis.
Seven annotators judged whether each record was natural and coherent in a blind review setting, without knowledge of the text's source.
Among the 128 records from \privasis, 113 were judged natural and coherent, compared to 111 from the human-written datasets.
This indicates that records in \privasis achieve a level of naturalness and coherence comparable to human-written records.

\begin{table}[t!]
\caption{
Diversity of \privasis domain subsets and human-written datasets.
}
\centering
\scriptsize
\renewcommand{\arraystretch}{1}
\begin{adjustbox}{width=0.87\linewidth}
\begin{tabular}{lcccc}
\toprule
\multirow{1}{*}{\textbf{Dataset}}
& \makecell{MATTR ($\uparrow$)} &
  \makecell{Bigram\\Diversity ($\uparrow$)} &
  \makecell{Shannon\\Entropy ($\uparrow$)} &
  \makecell{Cosine\\Similarity ($\downarrow$)} \\
\midrule
MIMIC-III Notes \citep{johnson2016mimic}
& 0.757 & \textbf{0.900} & 6.396 & 0.654 \\
\privasis Health \& Wellness
& \textbf{0.815} & 0.872 & \textbf{7.402} & \textbf{0.321} \\
\cmidrule{1-1}
GovReport \citep{huang-etal-2021-efficient}
& 0.781 & 0.813 & 7.071 & 0.354 \\
\privasis Government \& Civic
& \textbf{0.815} & \textbf{0.865} & \textbf{7.411} & \textbf{0.347} \\
\cmidrule{1-1}
Enron Email \citep{klimt2004enron}
& 0.794 & \textbf{0.897} & 5.871 & 0.331 \\
\privasis Media \& Comms.
& \textbf{0.823} & 0.877 & \textbf{7.448} & \textbf{0.320} \\
\cmidrule{1-1}
Finance Tasks \citep{cheng2024financetasks}
& 0.700 & 0.566 & 5.729 & 0.679 \\
\privasis Business \& Finance
& \textbf{0.807} & \textbf{0.864} & \textbf{7.346} & \textbf{0.353} \\
\cmidrule{1-1}
TAB \citep{pilan-etal-2022-text}
& 0.741  & 0.747 & 7.065 & 0.670 \\
\privasis Legal \& Compliance
& \textbf{0.817}  & \textbf{0.863} & \textbf{7.488} & \textbf{0.352} \\
\bottomrule
\end{tabular}
\end{adjustbox}
\label{tab:diversity}
\vspace{-10pt}
\end{table}

\noindent\textbf{Do the generated profiles correspond to real people?}
We sample 100 profiles to investigate whether they correspond to memorized real-world data.
Using Gemini-2.5-Pro Deep Research, we check if each profile matched a real person, providing full details (name, age, sex, citizenship, email, URLs, phone).
In case of URLs, none of them were accessible.
Of the 100 profiles, 5 were incomplete due to off-topic responses. In 15 cases, the model returned multiple potential matches, which we manually disambiguated: 8 shared the exact name and 7 had similar names, but other attributes (sex, age, nationality) did not align, and no email addresses matched.
In 3 cases, the model reported an exact match (name, sex, nationality), but manual review showed major discrepancies in age and contact information.
The remaining profiles were all reported as fabricated.
We also run a larger check on 1K profiles with websearch-enabled GPT-5 and none of them were judged to be real.
Thus, the profiles are synthetic and do not represent real individuals, reducing privacy concerns.

\section{Building a Sanitization Parallel Corpus}
\label{sec:sanitization}

As one concrete downstream application that leverages \privasis, we focus on training a sanitization model that can selectively remove or abstract sensitive information while preserving coherence and utility.
We aim to train a model that meets the following four goals:
(1) process diverse text domains,
(2) follow arbitrary sanitization instructions rather than being limited to fixed personally identifiable information (PII) categories,
(3) support multiple levels of abstraction beyond simple masking or deletion, and
(4) remain lightweight enough for local deployment.

Using \privasis, we build \sanitizationdataset, a high-quality sanitization corpus of triplets (original record, instruction, sanitized record) that treats sensitivity as contextual and supports flexible strategies that balance privacy with utility.
It is desirable for sanitization models to be small enough to run locally, so that sensitive text never needs to leave a user's device.
However, we find that even frontier LLMs struggle with sanitizing long text records effectively (\S \ref{subsec:exp_results}).
To address this challenge, we introduce a decomposition-based pipeline that breaks records into manageable chunks, enabling grounded and consistent sanitization.

\subsection{Sanitization Pipeline}
\label{subsec:sanitization_pipeline}

\noindent\textbf{1. Decomposition:}  
To make sanitization tractable, the record $x$ is recursively split into a set of chunks $\mathcal{C} = \{c_1, c_2, \dots\}$ using double newlines, EOS markers, or other natural boundaries until each chunk satisfies $|c_i| \leq \tau$, where $\tau = 512$ characters.  
This variable-length decomposition simplifies the sanitization task while preserving local coherence (e.g., list placed in the same chunk).  

\noindent\textbf{2. Target Selection:}  
From the annotated attributes $\mathcal{A}$ of $x$, 
we assign a sensitivity weight $w_a$ to each $a \in \mathcal{A}$ using an LLM, prioritizing highly sensitive information over relatively benign details that are difficult to sanitize (e.g., happy emotion).
Next, using these weights, we sample a set of $n$ targets, denoted $\mathcal{T} = \{z_1, \dots, z_n\}$, which may be individual attributes or attribute groups (\S \ref{subsec:synthesis_pipeline}).  
Each $z$ is then randomly labeled with $\ell_z \in \{\textsc{abstract}, \textsc{drop}\}$.  
By selecting targets stochastically, we go beyond PIIs to cover various information that users may contextually consider sensitive.

\noindent\textbf{3. Sanitization:}  
(i) For each target $z \in \mathcal{T}$, we first identify relevant chunks $\mathcal{C}_z \subseteq \mathcal{C}$ using an LLM.  
(ii) From each $c \in \mathcal{C}_z$, we extract the spans $\mathcal{S}_{z,c}$ that correspond to $z$.  
(iii) We then build a sanitization instruction $\mathsf{instr}_z$ for $z$:
\hspace{0.5em}If $\ell_z = \textsc{abstract}$, all $c \in \mathcal{C}_z$ are concatenated and passed to the LLM to generate an abstraction instruction grounded in all the relevant context (e.g., ``\textit{Abstract the specific date as `in the coming months'}'').
\hspace{0.5em}If $\ell_z = \textsc{drop}$, we use a fixed instruction (e.g., ``\textit{Drop the information about \{$z$\} from the text}'').
(iv) We use $\mathsf{instr}_z$ to sanitize each $c \in \mathcal{C}_z$ for consistency.
This decomposition-based strategy ensures both consistent abstraction across chunks and improved efficiency, since sanitization can be applied to chunks in parallel.  
Finally, we merge the sanitized chunks to reconstruct the sanitized record $\tilde{x}$.  
Further details are in Algorithm~\ref{alg:sanitization} in the Appendix.

\noindent\textbf{4. Final Instruction Generation:}  
After sanitization, we prompt an LLM to generate a final coherent instruction $\widehat{\mathcal{I}}$ based on all $\{\mathsf{instr}_z\}_{z \in \mathcal{T}}$.
To support scenarios where utility is important, we additionally include a set of \textit{retention target attributes} $\mathcal{K} = \{k_1, \dots, k_m\} \subseteq \mathcal{A}$, representing information that should be explicitly retained.  
We select $\mathcal{K}$ to minimize interference with the sanitization process, choosing attributes with the lowest lexical overlap with the sanitization targets $\mathcal{T}$, based on ROUGE scores.
When the \textit{grouped attributes} (\S \ref{subsec:synthesis_pipeline}) are selected as targets,
we occasionally omit the individual attribute names and only include the group label in $\widehat{\mathcal{I}}$, instead of iterating over the individual attributes. 
For example, 
\textit{``Please abstract all information related to locations while keeping the city.''} rather than \textit{``Please abstract the clinic address, session room, and the patient's address while keeping the city.''}
This encourages contextual generalization to natural user requests.
Finally, the pipeline yields a triplet $(\text{record } x, \text{instruction } \widehat{\mathcal{I}}, \text{sanitized record } \tilde{x})$, which supports instruction-following for sanitization.
More details of our pipeline are provided in Appendix \ref{app:sanitization}.

\subsection{Statistics \& Analysis}
\noindent\textbf{Basic Statistics:}
We construct a dataset of 100K examples and use 37K of them to build our training set, with 70.4\% generated by GPT-OSS-120B \citep{openai2025gptoss} and 29.6\% by Qwen3-80B \citep{yang2025qwen3}.
Sanitization instructions average 69.4 words and specify 2.9 targets. 
We construct 2.1K evaluation records using latest frontier models (GPT-5, Gemini-2.5-Pro, Qwen3-235B, LLaMA-4 Maverick). 
Further evaluation details are in Section~\ref{subsec:evaluation}.

\noindent\textbf{Comparison with Existing Privacy Datasets:}
Table~\ref{tab:privacy-datasets} compares \sanitizationdataset with existing datasets (extended analysis in Appendix~\ref{app:ext-related}). 
Most prior work focuses on PII span detection~\citep{douglass2004computer,pilan-etal-2022-text,papadopoulou-etal-2022-neural,gretel2024finance,zeng2025privacyannotation}, providing only deletion without rewritten alternatives, restricting to predefined PII categories, and operating in single domains with short text units.
\privasis records average 527 words, significantly longer than existing datasets, with annotated spans averaging 35.3 characters. 
The Self-Disclosure Corpus \citep{dou-etal-2024-reducing} spans average 28.7 characters but provides only single-level abstraction for pre-specified categories in narrow domains like Reddit posts.
In contrast, \privasis uniquely provides multiple abstraction levels with flexible, instruction-based sanitization supporting unlimited user-specified categories. 
Our dataset enables both removal and graded abstraction (e.g., ``March 3rd, 2024'' → ``early March'' → ``this spring'' → ``recently''), covers 100K+ multi-domain records, and generates natural language instructions for arbitrary privacy requirements.

\begin{table}[t!] \centering \small \setlength{\tabcolsep}{6pt} \renewcommand{\arraystretch}{1.15} 
\caption{
Comparison of public privacy datasets and related resources.
} 
\vspace{-5pt}
\label{tab:privacy-datasets} 
\begin{adjustbox}{max width=\textwidth} 
\begin{tabular}{@{}l r c c c c c c l@{}} 
\toprule 
\textbf{Dataset} & \textbf{Size} & \makecell{\textbf{Length}\\\textbf{(\#Words)}} & \makecell{\textbf{Sens.}\\\textbf{Spans}} & \makecell{\textbf{PII}\\\textbf{Spans}} & \makecell{\textbf{Abstr.}\\\textbf{Pairs}} & \textbf{Synthetic} & \textbf{Public} & \textbf{Domain} \\ 
\midrule 
MIMIC-II De-identification~\citep{douglass2004computer} & 2M & 282 & \xmark & \cmark & \xmark & \xmark & \xmark & Clinical \\ 
Text Anonymization Benchmark~\citep{pilan-etal-2022-text} & 1.2K & 843 & \cmark & \cmark & \xmark & \xmark & \cmark & Legal \\ 
Self-Disclosure Corpus~\citep{dou-etal-2024-reducing} & 4.8K & 29 & \cmark & \xmark & \cmark & \xmark & \cmark & Reddit \\ 
Gretel Synthetic Financial PII~\citep{gretel2024finance} & 55.9K & 188 & \xmark & \cmark & \xmark & \cmark & \cmark & Finance \\ 
SynthPAI~\citep{yukhymenko2024synthpai} & 7.8K & 17 & \xmark & \xmark & \xmark & \cmark & \cmark & Reddit \\ 
NaP$^2$ \citep{huang-etal-2025-nap2} & 4.8K & 171 & \xmark & \cmark & \cmark & \cmark & \cmark & Persona Chat \\ 
PANORAMA~\citep{selvam2025panorama} & 384K & 48 & \xmark & \xmark & \xmark & \cmark & \cmark & Multi \\ 
Automated Privacy Info Annotation~\citep{zeng2025privacyannotation} & 154K & 226 & \cmark & \cmark & \xmark & \xmark & \cmark & LLM queries \\ 
PAPILLON / PUPA~\citep{li-etal-2025-papillon} & 0.9K & 181 & \xmark & \xmark & \xmark & \xmark & \cmark & Dialogues \\ 
SemSI-Set / SemSI-Bench~\citep{zhang2025semsi} & 10.8K & 210 & \xmark & \xmark & \xmark & \cmark & \cmark & News \\ 
\rowcolor{green!10} \textbf{\sanitizationdataset (Ours)} & 1.4M(100K) & 527 & \textbf{\cmark} & \textbf{\cmark} & \textbf{\cmark} & \textbf{\cmark} & \textbf{\cmark} & Multi \\ 
\bottomrule 
\end{tabular} 
\end{adjustbox} 
\end{table}

\section{Experiments}
\label{sec:experiments}

\subsection{Model Training}

We train a sanitizer \sanitizer that, given text $x$ and instruction $\widehat{\mathcal{I}}$, outputs a sanitized version $\tilde{x}$ where target attributes are abstracted or removed.
Since it is safer when sanitization models are run locally, we target lightweight Qwen3 models: 0.6B, 1.7B, and 4B. We train them on a 37K subset of \sanitizationdataset (\S \ref{sec:sanitization}). 
Further details are in Appendix \ref{app:training}.

\subsection{Hierarchical Evaluation}
\label{subsec:evaluation}

We evaluate models on the \sanitizationdataset test set, which contains records from four frontier models: Gemini-2.5-pro, GPT-5, Llama-4-Maverick, and Qwen3-235B.
To assess sanitization effectiveness, we use a hierarchical evaluation framework capturing three information leakage types in sanitized text: (1) direct leak, (2) inference leak, and (3) proximity leak.
Our evaluation setup relies on exact string matching and factual questions to minimize potential evaluator bias.
The LLM evaluator is used to answer those factual questions, instead of preference based judgments.

First, we check for \textbf{direct leak} by performing exact string matching to determine whether the target attribute value $z_{value}$ appears verbatim in the sanitized record.
If absent, we test \textbf{inference leak} by prompting an evaluator LLM (GPT-OSS-120B) to predict the attribute value given only the sanitized text and attribute key $z_{key}$ (e.g., ``Please guess the name of the journal. Make a guess even if it's not included in the given text.'').
We denote this prediction as $\hat{z}_{sanitized}$ and check for exact string matches with $z_{value}$.
If no match occurs, we check \textbf{proximity leak} by comparing the evaluator's predictions from both the sanitized text ($\hat{z}_{sanitized}$) and the original record ($\hat{z}_{original}$).
The evaluator assesses which prediction is closer to the true attribute value $z_{value}$; if $\hat{z}_{sanitized}$ is as close as, or closer than, $\hat{z}_{original}$, this indicates a proximity leak and thus a sanitization failure.  
The sanitization of a record succeeds only if no target attributes $\mathcal{T}$ exhibit leakage (the \textit{Successful Record} metric).
This approach captures multiple levels of leakage from explicit disclosure to subtle inference.

Since a model returning an empty string would avoid all leakage, we also measure information retention. 
For each record, we check whether the retention target attributes ($\mathcal{K}$; \S \ref{subsec:sanitization_pipeline}) remain in the sanitized record using exact string matching and LLM. 
A record is considered successfully processed only if no information leakage occurs for any sanitization target and all retention target attributes are preserved in the sanitized text (the \textit{Full Successful Record} metric).
We also report \textit{Successful Attribute} and \textit{Successful Att./Record} metrics, which indicate the overall success rate across all attributes in the test set and the average success rate of attributes within each record, respectively.

We release two test sets: (1) \textbf{Vanilla} and (2) \textbf{Hard}.
The vanilla set (1,042 records) consists of records on which our sanitization pipeline (\S\ref{sec:sanitization}) achieves a perfect Full Successful Record score, while the hard set (1,149 records) contains records where even our decomposition-based pipeline fails to do so.
The difference lies mainly in grouped attributes: 60\% in vanilla vs.\ 87\% in hard, which require contextual target identification, thereby adding an extra layer of complexity.
Hard-set records are also longer (619.6 vs.\ 569.3 words) and paired with longer instructions (94 vs.\ 57.2 words), reflecting higher complexity.
Further details and examples of each leakage type are in Appendix \ref{app:evaluation} and \ref{app:error_analysis}.

\subsection{Results on \sanitizationdataset}
\label{subsec:exp_results}

\noindent\textbf{Overall Performance:}
Table \ref{tab:eval_results} shows results on both the Vanilla and Hard test sets.

(1) Vanilla test set:
Sanitization is fundamentally a re-writing task, yet even the strongest frontier models fall short of perfect performance, with Full Success Record rates ranging from 64.4\% (Qwen3-235B) to 70.3\% (o3).
Note, the Vanilla Test set corresponds to a subset of \privasis where our decomposition-based sanitization pipeline with GPT-4.1 achieves a perfect score. 
This gap indicates that frontier models even with reasoning capabilities struggle to reliably execute fine-grained sanitization.
In contrast, \sanitizer-4B, trained on \privasis, achieves 72.50\%, outperforming all frontier LLMs despite being orders of magnitude smaller.
Even \sanitizer-0.6B model, outperforms GPT-OSS-120B, Llama-4 Maverick, and Qwen3-235B, while the base models Qwen3 4B and 0.6B lags at 53.65\% and 16.70\%, respectively. 

\begin{table}[t!]
\vspace{-10pt}
\caption{
Sanitization performance of off-the-shelf LLMs and our \sanitizer models.
`Att.' denotes attribute and `Successful Att. / Record' denotes the average success rate within a record.
}
\centering
\scriptsize
\renewcommand{\arraystretch}{1.07}
\begin{adjustbox}{width=\linewidth}
\begin{tabular}{lccccccc}
\toprule
\multirow{2}{*}{\textbf{Model}} &
\multicolumn{3}{c}{\textbf{Sanitization}} &
\multicolumn{3}{c}{\textbf{Retention}} &
\makecell{\textbf{Full}} \\
\cmidrule(l{0.25em}r{0.25em}){2-4} \cmidrule(l{0.25em}r{0.25em}){5-7} \cmidrule(l{0.25em}r{0.25em}){8-8}
& \makecell{Successful\\Attribute (\%)} &
  \makecell{Successful Att.\\/ Record (\%)} &
  \makecell{Successful\\Record (\%)} &
  \makecell{Successful\\Attribute (\%)} &
  \makecell{Successful Att.\\/ Record (\%)} &
  \makecell{Successful\\Record (\%)} &
  \makecell{Successful\\Record (\%)} \\
\midrule
\multicolumn{8}{l}{\textbf{Vanilla Test Set}} \\
o3 
& 91.53 & 91.62 & 74.57 
& 94.06 & 95.24 & 93.57 
& \underline{70.25} \\
DeepSeek R1 
& 88.54 & 91.01 & 72.26 
& 93.74 & 94.39 & 93.38 
& 69.58 \\
GPT-5 
& 90.16 & 91.80 & 73.90 
& 93.42 & 94.08 & 92.71 
& 70.06 \\
GPT-4.1 
& 88.07 & 90.73 & 72.26 
& 94.48 & 95.80 & 94.43 
& 70.06 \\
GPT-OSS-120B 
& 89.42 & 90.33 & 69.87
& 95.28 & 95.61 & 94.53 
& 67.66 \\
LLaMA-4 Maverick 
& 89.31 & 90.77 & 73.61 
& 89.97 & 90.27 & 88.68 
& 67.18 \\
Qwen3-235B 
& 83.34 & 87.58 & 68.04 
& 93.47 & 94.31 & 93.19 
& 64.40 \\
Qwen3-4B
& 75.66 & 79.59 & 57.58 
& 90.66 & 90.95 & 90.02 
& 53.65 \\
Qwen3-0.6B
& 47.39 & 53.77 & 35.99 
& 70.49 & 71.42 & 69.19 
& 16.70 \\
\cmidrule{1-1}
\sanitizer-4B
& 86.60 & 90.60 & 72.84 
& 99.15 & 99.13 & 98.66 
& \textbf{72.50} \\
\sanitizer-0.6B
& 81.61 & 87.37 & 68.52 
& 99.26 & 99.23 & 98.75 
& 68.04 \\
\midrule
\multicolumn{8}{l}{\textbf{Hard Test Set}} \\
o3 
& 80.20 & 75.65 & 15.23 
& 87.89 & 87.04 & 83.81 
& 11.66 \\
DeepSeek R1 
& 75.54 & 72.76 & 15.14 
& 86.70 & 86.16 & 82.59 
& 11.23 \\
GPT-5 
& 78.78 & 75.30 & 16.28 
& 87.08 & 87.23 & 84.25 
& \textbf{13.14} \\
GPT-4.1 
& 75.14 & 72.40 & 13.93 
& 89.79 & 90.06 & 86.51 
& 12.18 \\
GPT-OSS-120B 
& 77.67 & 74.07 & 13.84 
& 88.36 & 87.87 & 84.94 
& 10.53 \\
LLaMA-4 Maverick 
& 76.21 & 73.40 & 16.19 
& 82.07 & 81.92 & 78.24 
& 11.05 \\
Qwen3-235B 
& 69.01 & 67.33 & 12.79 
& 89.37 & 89.71 & 86.95 
& 10.27 \\
Qwen3-4B
& 62.15 & 59.89 & 8.27 
& 84.36 & 84.01 & 80.68 
& 5.66 \\
Qwen3-0.6B
& 33.41 & 35.46 & 12.53 
& 76.63 & 76.08 & 72.58 
& 4.44 \\
\cmidrule{1-1}
\sanitizer-4B
& 73.29 & 71.23 & 14.19 
& 95.76 & 95.60 & 92.96 
& \underline{12.36} \\
\sanitizer-0.6B
& 68.11 & 66.14 & 11.31 
& 95.09 & 95.20 & 91.99 
& 9.31 \\
\bottomrule
\end{tabular}
\end{adjustbox}
\label{tab:eval_results}
\end{table}

(2) Hard test set:
Performance declines sharply for all models on the Hard test set, which introduces more challenging attribute selections and longer contexts.
Frontier models drop to only 10–13\% Full Success, with GPT-5 reaching the highest at 13.14\%.
Our model again matches frontier-level performance, achieving 12.80\%, outperforming o3 and R1, ranking second only to GPT-5.
These results underscore both the challenge of fine-grained sanitization and the effectiveness of our dataset.  

\noindent\textbf{Sanitization Metrics \& Retention Metrics:}
Although the Successful Record scores for most models are around 70\%, their Successful Attribute/Record scores exceed 90\%.
This gap reveals a critical weakness: while models sanitize the majority of attributes, they routinely miss at least one per record.
For example, although the Successful Attribute score for our model is lower than o3 and same with GPT-5, the full success record rate of our model is higher.
Such failures are unacceptable in privacy-sensitive settings, because one missed attribute is enough to compromise the entire record, no matter how many others were sanitized correctly.
Frontier models also underperform on retention, frequently over-editing and altering non-target attributes.
For example, GPT-5 retains only 93.4\% of attributes on the Vanilla set, versus 99.2\% for \sanitizer-4B.  
These highlight a core weakness of frontier LLMs: reliably separating sensitive from non-sensitive information.

\subsection{Analysis}

\noindent\textbf{How robust is the LLM-based proximity leak evaluation?}
We run human evaluation on all proximity leak failure cases ($n=140$) produced by the o3 model and found 98\% of them to be correct, indicating the high precision of our evaluation setup. 
We additionally assessed robustness by repeating the proximity leak evaluation using two different models (GPT-OSS-120B and Qwen3-80B), achieving a high inter-model agreement of 97\% over 5K cases, indicating low susceptibility to evaluator-specific bias.

\noindent\textbf{What types of information leakage do the models make?}
Table \ref{tab:info_leakage} reports the information leakage patterns (\S \ref{subsec:evaluation}) of models on the Vanilla test set of \sanitizationdataset.
Strikingly, all models most frequently exhibit direct leaks, exposing sensitive information in verbatim form in the supposedly ``sanitized'' outputs.
Our \sanitizer-4B model achieves the third-lowest ratio of direct leak, following GPT-OSS-120B and o3.
In contrast, the base 4B model shows the highest ratio, indicating that training on our dataset improves information leakage patterns as well.
GPT-OSS-120B shows the lowest direct leak ratio, suggesting it can better identify the target attributes for sanitization; however, it ultimately fails to sanitize them effectively, as reflected by its Successful Record score in Table \ref{tab:eval_results}.
Interestingly, our \sanitizer-0.6B model shows the highest direct leak ratio, but outperforms Qwen3-235B on the Full Successful Record score.
Among the frontier models, Qwen3-235B stands out with a notably high direct leak ratio, which may reflect a broader limitation within the Qwen3 model family.

\begin{minipage}{0.37\textwidth}
Note that the ratio of inference leaks is significantly lower than proximity leaks because we apply exact string matching on top of the evaluator's predictions of the attribute (\S \ref{subsec:evaluation}).
The inference leak ratio would likely increase if semantic entailment were used instead, since some attributes (e.g., lists or long string) are difficult to capture via exact string matching.
To account for such cases, we rely on the proximity leak metric instead. 
Case studies are provided in Appendix \ref{app:error_analysis}.
\end{minipage}\hfill
\begin{minipage}{0.59\textwidth}

\begin{table}[H]
\caption{Ratio of each information leakage type (\S \ref{subsec:evaluation}).}
\vspace{-8pt}
\centering
\scriptsize
\renewcommand{\arraystretch}{1.02}
\begin{adjustbox}{width=0.98\linewidth}
\begin{tabular}{lccc}
\toprule
\multirow{1}{*}{\textbf{Model}}
& \makecell{Direct\\Leak} &
  \makecell{Inference\\Leak} &
  \makecell{Proximity\\Leak} \\
\midrule
o3
& 60.4\%  & 8.4\% & 31.2\% \\
DeepSeek R1
& 70.0\%  & 6.7\% & 23.3\% \\
GPT-5 
& 67.4\%  & 7.7\% & 24.9\% \\
GPT-OSS-120B 
& 53.7\%  & 8.4\% & 37.9\% \\
LLaMA-4 Maverick 
& 66.1\%  & 6.2\% & 27.7\% \\
Qwen3-235B 
& 75.3\%  & 4.3\% & 20.4\% \\
Qwen3-4B
& 81.7\%  & 3.1\% & 15.2\% \\
\cmidrule{1-1}
\sanitizer-4B
& 64.8\%  & 8.4\% & 26.8\% \\
\sanitizer-0.6B
& 79.4\%  & 3.1\% & 17.5\% \\
\bottomrule
\end{tabular}
\end{adjustbox}
\label{tab:info_leakage}
\end{table}

\vspace{5pt}
\end{minipage}

\noindent\textbf{Where do the models struggle most?}
Table \ref{tab:top_failed_categories} presents the top 8 main categories of records (\S \ref{subsec:privasis_analysis}) where the models fail most often. 
Overall, \textit{Business \& Finance} is the most challenging category, followed by \textit{Health \& Wellness}. 
The former primarily includes financial records, while the latter covers medical records. 
A notable pattern is that our \sanitizer model exhibits a more balanced performance across categories, whereas o3 struggles disproportionately with \textit{Health \& Wellness} compared to other domains.
Table \ref{tab:top_failed_attrs} in the Appendix summarizes the top 8 attributes that models fail most often.
We find the models struggle the most with name-related attributes (e.g., last name, full name, and  user handle), and dates.

\begin{table}[t]
\caption{Top 8 main categories where each model fails and their ratio.}
\centering
\scriptsize
\setlength{\tabcolsep}{3pt}
\renewcommand{\arraystretch}{1.02}
\begin{adjustbox}{width=0.98\linewidth}
\begin{tabular}{c*{8}{c}}
\toprule
\textbf{Model} & 1 & 2 & 3 & 4 & 5 & 6 & 7 & 8 \\
\midrule
o3 & \makecell{Health \& \\ Wellness \\ (26.5\%)} & \makecell{Business \\\& Finance \\ (17.1\%)} & \makecell{Legal \& \\ Compliance \\ (9.6\%)} & \makecell{Education \\\& Training \\ (9.1\%)} & \makecell{Personal \\\& Family \\ (9.1\%)} & \makecell{Government \\\& Civic \\ (7.8\%)} & \makecell{Community \\\& Social \\ (6.0\%)} & \makecell{Professional \\ Services \\ (6.0\%)}  \\
\midrule
DeepSeek R1 & \makecell{Business \\\& Finance \\ (22.7\%)} & \makecell{Health \& \\ Wellness \\ (19.2\%)} & \makecell{Government \\\& Civic \\ (12.0\%)} & \makecell{Personal \\\& Family \\ (9.9\%)} & \makecell{Education \\\& Training \\ (8.6\%)} & \makecell{Legal \& \\ Compliance \\ (8.1\%)} & \makecell{Professional \\ Services \\ (7.2\%)} & \makecell{Community \\\& Social \\ (4.9\%)}  \\
\midrule
GPT-5 & \makecell{Health \& \\ Wellness \\ (22.3\%)} & \makecell{Business \\\& Finance \\ (19.1\%)} & \makecell{Education \\\& Training \\ (11.0\%)} & \makecell{Legal \& \\ Compliance \\ (9.7\%)} & \makecell{Government \\\& Civic \\ (9.4\%)} & \makecell{Personal \\\& Family \\ (7.3\%)} & \makecell{Community \\\& Social \\ (5.3\%)} & \makecell{Media \& \\ Communications \\ (4.6\%)}  \\
\midrule
\makecell{LLaMA-4\\Maverick} & \makecell{Business \\\& Finance \\ (21.5\%)} & \makecell{Health \& \\ Wellness \\ (17.5\%)} & \makecell{Government \\\& Civic \\ (11.8\%)} & \makecell{Education \\\& Training \\ (8.8\%)} & \makecell{Legal \& \\ Compliance \\ (8.5\%)} & \makecell{Personal \\\& Family \\ (8.3\%)} & \makecell{Community \\\& Social \\ (6.9\%)} & \makecell{Professional \\ Services \\ (6.7\%)}\\
\midrule
Qwen3-235B & \makecell{Business \\\& Finance \\ (19.7\%)} & \makecell{Health \& \\ Wellness \\ (19.5\%)} & \makecell{Legal \& \\ Compliance \\ (11.2\%)} & \makecell{Personal \\\& Family \\ (11.2\%)} & \makecell{Education \\\& Training \\ (10.2\%)} & \makecell{Government \\\& Civic \\ (10.0\%)} & \makecell{Professional \\ Services \\ (5.8\%)} & \makecell{Community \\\& Social \\ (5.1\%)} \\
\midrule
\makecell{\privasis\\\textsc{Sanitizer}-4B} & \makecell{Business \\\& Finance \\ (15.2\%)} & \makecell{Education \\\& Training \\ (15.1\%)} & \makecell{Health \& \\ Wellness \\ (13.8\%)} & \makecell{Personal \\\& Family \\ (13.6\%)} & \makecell{Government \\\& Civic \\ (13.0\%)} & \makecell{Community \\\& Social \\ (11.1\%)} & \makecell{Legal \& \\ Compliance \\ (7.9\%)} & \makecell{Professional \\ Services \\ (6.8\%)}\\
\bottomrule
\end{tabular}
\end{adjustbox}
\vspace{-8pt}
\label{tab:top_failed_categories}
\end{table}

\noindent\textbf{How well does \sanitizer generalize?}
We evaluate the performance of \sanitizer-4B on the NaP² dataset \citep{huang-etal-2025-nap2}, which includes high-quality human-rewritten text for sanitization, in a zero-shot setting.
We follow the proximity-leak evaluation protocol using the sanitization target information (i.e., sensitive profile information) provided by NaP². 
Specifically, we prompt the evaluator model (GPT-OSS-120B) to judge whether the original text or the sanitized text is closer to the sensitive information. 
If the evaluator judges the sanitized text is as close as, or closer than, the original text, we consider that a leak.

The 4B model finetuned directly on NaP² achieves a low leak ratio of 10\%. 
Our \sanitizer 4B model, despite never being trained on NaP², achieves the same leak ratio of 10\%.
On the other hand, the NaP²-trained model scores 31.96\% on \privasis's Full Successful Record metric, which is significantly lower than \sanitizer-4B (72.5\%).
This demonstrates the strong robustness of the \sanitizer model and suggests that training on \privasis yields superior generalization due to its scale and diversity.

\begin{table}[t]
\centering
\footnotesize
\caption{
Zero-shot generalization performance of \sanitizer-4B on the NaP² dataset compared to the NaP²-finetuned model.
Note, \sanitizer-4B was not trained on NaP² and was evaluated in a zero-shot setting.
}
\label{tab:generalization-results}
\begin{tabular}{lcc}
\toprule
\textbf{Model} & \makecell{Leak Ratio \\on NaP² ($\downarrow$)} & \makecell{Full Successful Record \\on \privasis ($\uparrow$)} \\ 
\midrule
NaP²-Finetuned (4B) & 10.0 \% & 32.0 \% \\
\textbf{\sanitizer-4B (Ours)} & \textbf{10.0 \%} & \textbf{72.5 \%} \\
\bottomrule
\end{tabular}
\end{table}

\section{Related Work}
We position our work within privacy-preserving data generation. Table~\ref{tab:privacy-datasets} summarizes existing privacy datasets (extended discussion in Appendix~\ref{app:ext-related}).

\noindent\textbf{Synthetic Data Generation:} 
Related approaches include differential privacy methods using DP-SGD~\citep{abadi2016deep} or public-to-private pipelines~\citep{mattern-etal-2022-differentially,yue-etal-2023-synthetic,mckenna2025scaling,lin2024differentially,xie2024differentially,zhang2025pcevolve}. 
Non-private methods include self-instruction~\citep{wang-etal-2023-self-instruct,openr1,limo}, targeted prompting~\citep{textbooks-are-all-you-need,phi-4}, and automated refinement~\citep{advances_in_text_and_code, kim-etal-2023-soda,infosumm}, but rely on fixed prompts or seed data, limiting diversity~\citep{quality-diversity-complexity,prismatic-synthesis}. 
\privasis generates million-scale datasets from scratch using auxiliary control variables and diversity-preserving refinement without predefined prompts.

\noindent\textbf{PII Removal Datasets:} 
Classic corpora for anonymization establish span-level detection~\citep{stubbs2015i2b2,pilan-etal-2022-text}, while recent work expands through synthetic documents and LLM interactions~\citep{gretel2024finance,selvam2025panorama,zeng2025privacyannotation,yukhymenko2024synthpai}. 
These remain small-scale or domain-specific; \privasis provides large-scale, multi-domain coverage with both PII and sensitive spans.

\noindent\textbf{Data Minimization:}
Related work abstracts non-PII sensitive details through disclosure rewrites~\citep{dou-etal-2024-reducing}, naturalness benchmarks~\citep{huang-etal-2025-nap2}, LLM anonymization~\citep{staab2025anonymizers,zeng-etal-2025-privacyrestore}, and generalization strategies~\citep{olstad-etal-2023-generation,papadopoulou2023riskindicators}. 
\privasis unifies these approaches with fine-grained span labels and parallel abstraction/removal pairs across diverse document types.

\section{Conclusion}
We introduced \privasis, the first million-scale synthetic dataset with rich private information, addressing fundamental data scarcity in privacy-sensitive research.
Built from scratch using auxiliary control variables and diversity-preserving refinement, \privasis contains over 1M records spanning medical, financial, legal, email, calendar, and meeting domains.
Using this resource, we developed \sanitizationdataset with a decomposition-based pipeline enabling small models ($\leq$4B) to outperform frontier LLMs like GPT-5 and Qwen3-235B on text sanitization.%
We plan to release all code, data, and models, to stimulate research in privacy-preserving generation, controllable sanitization, and agentic systems processing sensitive data.

\section*{Ethics Statement}

The primary goal of our work is to address a fundamental bottleneck in privacy-sensitive research: the scarcity of large-scale public datasets containing rich sensitive attributes.

\paragraph{Methodology and Privacy Safeguards}
To bridge this gap, all data in the \privasis dataset are fully synthetic and were generated entirely from scratch for research and evaluation purposes only. 
The data generation process relies on auxiliary control variables and publicly available name databases and does not incorporate or reference any real-world private data. 
By construction, this approach avoids the ethical and legal risks associated with collecting, storing, or distributing sensitive human information.

All personal identifiers—including names, addresses, and Social Security numbers—are entirely fictitious, and any resemblance to real persons (living or deceased), business entities, or locations is purely coincidental. Manual verification procedures confirmed that no generated profiles correspond to actual individuals, ensuring that the dataset poses a low risk of privacy infringement or re-identification.

\paragraph{Intended Use and Real-World Validity}
The synthesis pipeline and dataset were designed explicitly to advance research in privacy-preserving machine learning, including the development and evaluation of data sanitization methods, differential privacy techniques, and responsible AI agent frameworks.
The dataset does not represent real-world events and should not be used for clinical decision-making, financial analysis, or as a basis for any action involving real individuals.

\paragraph{Community Commitment}
In alignment with the principles of transparency, reproducibility, and open science, we commit to releasing all models, code, and data to the research community.
Use of the dataset is restricted to non-commercial research and evaluation purposes.
We explicitly prohibit any attempts to re-identify individuals or to misuse the data for fraudulent or harmful activities.
This synthetic data framework enables ethical exploration of methods designed for privacy-critical settings while preserving human privacy by design.

\bibliography{papers, anthology-1, anthology-2}
\bibliographystyle{colm2026_conference}

\newpage
\appendix
\renewcommand\thesection{\Alph{section}}

\section{Synthesizing \privasis}
\label{app:synthesis}

\noindent\textbf{Generating Profiles:}
Given a sampled first name from the US SSN applicant database, we prompt the LLM with a fixed set of attributes to fill it: last name, sex, ethnicity, citizenship, ID type, ID number, passport number, phone number, email, user handle, URL, and life event.
For sex, ethnicity and life event, we provide predefined options for the LLM to choose.
Box \ref{box:profile_template} shows the template.

\begin{mybox}[label=box:profile_template]{Prompt for Generating Profiles}
\scriptsize

I will provide a set of demographic attributes of a person. Generate a complete personal profile by populating the fields in the provided structure below. Ensure the entries for the profile fields are realistic, and consistent. Be creative. You will be given an option list for some attribute types to choose from for events and populate.
\\

\textbf{Demographic attributes} \\
\{\}\\
\\
Generate the following:

- Last Name: \\
- Sex: Choose from \{\} \\
- Ethnicity: Choose from \{\} \\
- Citizenship: \\

- ID type: \\
- ID Number: \\
- Passport Number: \\

- Phone Number: \\
- Email: \\
- User Handle: \\
- URL: \\

\textbf{Life event list} \\
\{\} \\
Or you can come up with attributes and details as you please.\\
\\
Now populate the above fields, include the provided demographic information in the output as well. Make necessary changes to ensure consistency. List event attributes under the key attributes. Only output the completed profile.
\end{mybox}

\noindent\textbf{Generating Record Types:}
Given a profile, we generate a list of candidate record types and randomly select one of them to promote diversity.
The prompt used is presented in Box \ref{box:record_type}.

\begin{mybox}[label=box:record_type]{Prompt for Generating Record Types}
\scriptsize

Profile:\\
\{profile\}\\
\\
Generate a diverse and realistic list of types of \{record formality\} that contains \{name\}'s private information '{attribute}'. \\
\\
Requirements:\\
1. Each record type must specify the exact source/organization (e.g., "Reddit post from r/relationships", "Patient record from Mayo Clinic", "Tax document from IRS", "Employment record from Google")\\
2. Include a mix of different contexts (personal, professional, medical, legal, financial, etc.)\\
3. Consider both digital and physical record types\\
4. Include both common and unusual/unique record types\\
5. Ensure all types are textual (no images/videos)\\
6. Each record type should be specific and detailed, not generic\\
\\
Output the list as an ordered list, with each type being specific and detailed. Do not include any additional comments.

\end{mybox}

\noindent\textbf{Generating Background Contexts:}
Given the profile and record type, we generate a list of background contexts and randomly select one of them to encourage diversity.
Box \ref{box:background_context} contains the prompt that we use.

\begin{mybox}[label=box:background_context]{Prompt for Generating Background Contexts}
\scriptsize

Profile:\\
\{profile\}\\
\\
Generate five creative and specific contexts for the '\{record type\}' that contains \{profile['first\_name']\}'s '\{attribute\}'.\\
\\
Requirements: Each context should be specific, detailed, realistic and plausible. Include a diverse range of emotional contexts. Consider unique different scenarios, and be detailed enough to understand the situation.\\
\\
Output the five contexts in an ordered list with each item in plain text. Each context should be specific and detailed, providing a clear situation without generating the actual record. Do not include any additional comments.\\

\end{mybox}

\noindent\textbf{Generating Record Formats:}
Given a record type and the background context, we generate a list of candidate record formats and randomly select one of them to promote diversity.
The prompt used is in Box \ref{box:record_format}.

\begin{mybox}[label=box:record_format]{Prompt for Generating Record Formats}
\scriptsize
Record type: \{record type\}\\
\\
Situation: \{background context\}\\
\\
Based on the situation described above, outline what the structure should look like for '\{record type\}'. Describe ten diverse and realistic possible structures and their tone for the '\{record type\}' written in plain text in an ordered list. Do not include any values, just a plain description of the structure and tone. Tone should be realistic and diverse, not too cheerful.
\end{mybox}

\noindent\textbf{Generating Records:}
Given the profile, record type, background context, and format, we generate the record with the prompt in Box \ref{box:record}.

\begin{mybox}[label=box:record]{Prompt for Generating Records}
\scriptsize
Generate a realistic, detailed and creative '\{record type\}' in English according to the situation above. Follow these guidelines:\\
\\
1. Use Profile Information:\\
   - Incorporate relevant attributes from \{name\}'s profile\\
   - Ensure all personal details match the profile exactly\\
\\
2. Add Specific Details:\\
   - Include exact dates (avoid using 15th, use other random dates)\\
   - Specify precise locations with addresses or landmarks\\
   - Add realistic timestamps and durations\\
   - Include specific measurements, quantities, and numbers\\
   - Instead of monotonic numbers (e.g., 12345, 9876), use other random numbers\\
   - Use accurate terminology and jargon\\
\\
3. Structure and Format:\\
   - Follow the specified structure and tone: \{form\}\\
   - Keep the text dense and information-rich\\
   - Minimize markdown formatting\\
\\
Only output the generated '\{record type\}' without any additional comments or explanations.\\

\end{mybox}

\noindent\textbf{Cost for API models:}
Generating 10K records with GPT-4.1 costs approximately \$900, while sanitizing 10K records with GPT-4.1 costs about \$1,100.

\subsection{Analysis}
\label{app:subsec:privasis_analysis}

\noindent\textbf{Comparing Different LLMs for the Synthesis Pipeline:}

We compare multiple models using diversity metrics and output length, and we also performed manual quality inspection.
Table \ref{tab:generator_comparison} summarizes the results.
Although some models outperform GPT-OSS-120B on individual metrics, its performance is consistently strong across all measures.
It also generates longer text than other models, which is crucial for synthesizing documents.
A manual review also showed that GPT-OSS-120B generations have high coherence. 
When factoring in cost, it provides a substantially better price-performance ratio than frontier models (e.g., Gemini-2.5-pro, GPT-5, Qwen3-235B), making it more suitable for large-scale generation.

\begin{table}[h!]
\caption{Model diversity and lexical metrics.}
\centering
\begin{adjustbox}{width=\linewidth}
\begin{tabular}{lcccccc}
\toprule
Model & MATTR & Bigram Diversity & Shannon Entropy & Cosine Similarity & Vendi Score & Number of Words \\
\midrule
gemini-2.5-pro   & 0.8031 & 0.9033 & 7.1188 & 0.2729 & 74.57 & 521.9 \\
gpt-5            & 0.8072 & 0.9076 & 7.3514 & 0.3287 & 50.90 & 1168.1 \\
gpt-4.1-mini     & 0.8332 & 0.8994 & 7.2104 & 0.2733 & 76.74 & 477.6 \\
gpt-oss-120b     & 0.8108 & 0.8970 & 7.2613 & 0.3188 & 62.75 & 611.8 \\
qwen3 235b       & 0.8278 & 0.9135 & 7.3384 & 0.2880 & 70.06 & 689.1 \\
qwen3-80b        & 0.8133 & 0.9162 & 7.1980 & 0.3249 & 58.27 & 463.2 \\
llama4-maverick  & 0.7988 & 0.8669 & 6.9019 & 0.2667 & 80.07 & 356.1 \\
llama-3.3-70b    & 0.7935 & 0.8665 & 6.8900 & 0.2817 & 74.58 & 422.1 \\
exaone3.5-32b    & 0.8036 & 0.8744 & 7.0504 & 0.2931 & 70.58 & 440.7 \\
\bottomrule
\end{tabular}
\end{adjustbox}
\label{tab:generator_comparison}
\end{table}

\noindent\textbf{Ablation on the Synthesis Pipeline:}
We conduct an ablation study on the components of our synthesis pipeline using GPT-OSS-120B to generate 500 records.
Table \ref{tab:synthesis_ablation} reports the diversity metrics (as in Table \ref{tab:diversity}) when specific components are removed.
We additionally report the Vendi score \citep{friedman2023vendi}.
Without the auxiliary control variables for record type and format, the semantic diversity of the synthesized records—measured by cosine similarity and the Vendi score—drops sharply, even though the diversity-preserving term is still present.
Similarly, when records are iteratively revised without the diversity-preserving term, semantic diversity decreases markedly.
In contrast, our full pipeline with diversity-preserving iterative refinement (\S \ref{subsec:synthesis_pipeline}) achieves significantly higher semantic diversity compared to the variant without revision.

\begin{table}[t!]
\vspace{-10pt}
\caption{
Diversity of ablation of the components in the \privasis synthesis pipeline.
}
\centering
\small
\renewcommand{\arraystretch}{1}
\begin{adjustbox}{width=0.9\linewidth}
\begin{tabular}{lccccc}
\toprule
\multirow{1}{*}{\textbf{Dataset}}
& \makecell{MATTR ($\uparrow$)} &
  \makecell{Bigram\\Diversity ($\uparrow$)} &
  \makecell{Shannon\\Entropy ($\uparrow$)} &
  \makecell{Cosine\\Similarity ($\downarrow$)} &
  \makecell{Vendi\\Score ($\uparrow$)}\\
\midrule
\privasis \\
\hspace{1em}without Record Type and Format
& 0.808 & \textbf{0.905} & \textbf{7.48} & 0.408 & 39.4 \\
\hspace{1em}without Revision
& \underline{0.809} & 0.885 & 7.07 & \underline{0.324} & \underline{53.2} \\
\hspace{1em}without Diversity-preserving Term
& 0.805 & 0.863 & \underline{7.40} & 0.354 & 50.1 \\
\hspace{1em}Full Pipeline
& \textbf{0.810} & \underline{0.890} & 7.14 & \textbf{0.322} & \textbf{57.3} \\
\bottomrule
\end{tabular}
\end{adjustbox}
\label{tab:synthesis_ablation}
\end{table}

\begin{figure}[h!]
    \centering
    \includegraphics[width=\linewidth]{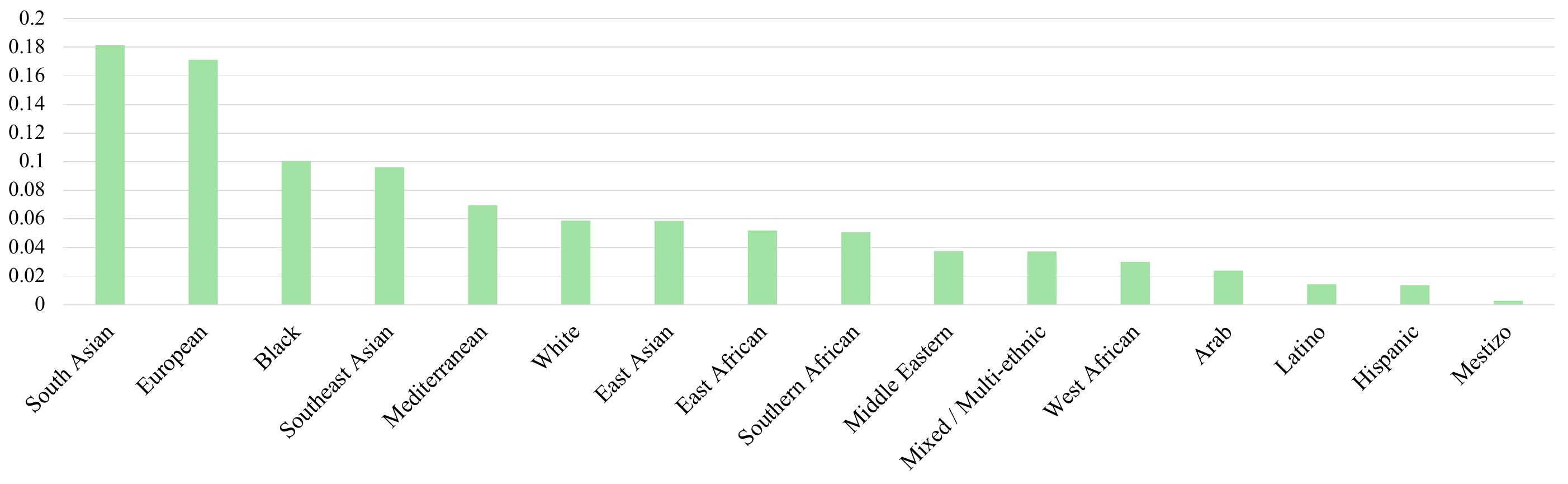}
    \caption{
    The ethnicity distribution in \privasis.
    } 
    \label{fig:ethnicity}
\end{figure}

\noindent\textbf{Distribution of the Generated Profiles:} Figure \ref{fig:ethnicity} shows the distribution of the ethnicity in the generated profiles.
We find that East Asian, Latin American, Hispanic, and Mestizo groups are particularly under-represented in \privasis, compared to real-world demographics.
Future work should aim to incorporate more balanced sampling to better reflect global demographics.

\noindent\textbf{Sanity Check on Profile Search with Authors:} As an additional sanity check, we ran Gemini-2.5-Pro Deep Research on all of the authors of this paper.
The model correctly confirmed all of us as real individuals, which increases confidence in its ability to distinguish fabricated from real profiles.

\noindent\textbf{Category Annotation:}
Box \ref{box:category_annotation_prompt} shows the prompt for labeling the broader categories for each record.
\begin{mybox}[label=box:category_annotation_prompt]{Category Annotation Prompt}
\scriptsize
Given this document, either select the most appropriate category from the existing list OR generate a new BROAD category name if none fit well.\\
\\
Existing categories: \{categories list\}\\
\\
Document: \{record\}\\
\\
Instructions:\\
- PREFER using an existing category if the document reasonably fits\\
- Only create a new category if the document is fundamentally different from existing ones\\
- New categories should be BROAD and GENERAL (like "financial records", "employment documents", "legal contracts")\\
- Avoid overly specific categories\\
- Respond with only the category name, nothing else. Do not include any other text or explanation.\\
\\
Category Name:
\end{mybox}

\begin{figure}[ht]
    \centering
    \includegraphics[width=\linewidth]{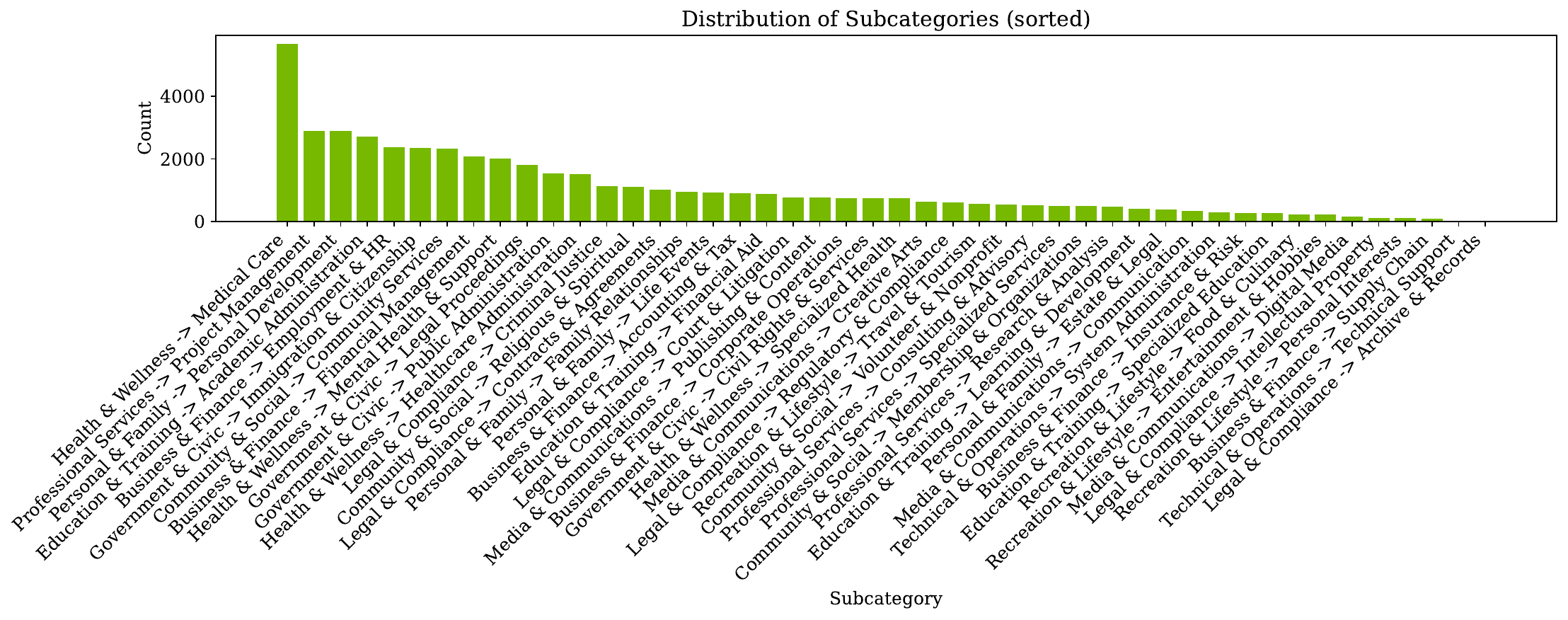}
    \caption{Distribution of the subcategories of the records in \sanitizationdataset.} 
    \label{fig:sanitization_data_subcategories}
\end{figure}

Figure \ref{fig:sanitization_data_subcategories} illustrates the distribution of the record subcategories.
The subcategory `Medical Care' is the most prevalent, comprising a significant portion of the broader main category \textit{Health \& Wellness}.

\noindent\textbf{Support for Synthesizing Multilingual Outputs:}
We also recognize the importance of generating non–English data. To this end, we leveraged non-U.S. LLMs in our data generation process, including Exaone 3.5 (South Korea) and Qwen3 (China).
Although the primary languages of these models are not English, they perform well within our English-language pipeline.
Therefore, we expect them to perform as good as when they are prompted in their major language.

To validate this, we generated Chinese and Korean records with Exaone 3.5 32B, Qwen3 80B, and GPT-4.1 through our pipeline.
We then conducted native-speaker human evaluation with both Korean and Chinese reviewers.
The reviewers confirmed that the generated records were coherent, contextually appropriate, and diverse (with cosine similarity scores of 0.34 and 0.36, respectively), making them comparable to English ones.
These results suggest that our synthesis pipeline can be directly applied to LLMs whose primary language is not English.

\section{Building the Sanitization Parallel Corpus}
\label{app:sanitization}

Algorithm \ref{alg:sanitization} describes our sanitization pipeline (\S \ref{subsec:sanitization_pipeline}) in detail.

\begin{algorithm}[ht]
\small
\caption{Sanitization Pipeline}
\label{alg:sanitization}
\begin{algorithmic}[1]
\Require Record $x$ with attributes $\mathcal{A}$; chunk threshold $\tau=512$ chars; number of targets $n$
\Ensure Sanitized record $\tilde{x}$ and user-style instruction $\widehat{\mathcal{I}}$
\Function{SanitizeRecord}{$x, \mathcal{A}, \tau, n$}
    \Statex \Comment{\textbf{1) Decomposition}}
    \State $\mathcal{C} \gets$ \Call{Decompose}{$x, \tau$} \Comment{Split $x$ until each chunk $|c|\le\tau$}
    \Statex
    \Statex \Comment{\textbf{2) Target Selection}}
    \State $W \gets$ \Call{SensitivityWeightLLM}{$\mathcal{A}$} \Comment{Higher $w_a$ for highly sensitive attributes}
    \State $\mathcal{T} \gets$ \Call{SampleTargets}{$\mathcal{A}, n, W$} \Comment{Targets may be individual attributes or attribute groups}
    \ForAll{$z \in \mathcal{T}$}
        \State $\ell_z \gets$ \Call{RandomLabel}{$\{\text{abstract}, \text{drop}\}$}
    \EndFor
    \Statex
    \Statex \Comment{\textbf{3) Sanitization}}
    \ForAll{$z \in \mathcal{T}$} \Comment{Chunk ops run in parallel within the same $z$, but run sequentially across different $z$}
        \State $\mathcal{C}_z \gets$ \Call{FindRelevantChunksLLM}{$z, \mathcal{C}$}
        \ForAll{$c \in \mathcal{C}_z$}
            \State $\mathcal{S}_{z,c} \gets$ \Call{ExtractSpansLLM}{$z, c$}
        \EndFor
        \If{$\ell_z = \textsc{abstract}$}
            \State $\mathsf{instr}_z \gets$ \Call{BuildAbstractionInstrLLM}{$z, \textstyle\bigcup_{c \in \mathcal{C}_z} c$}
        \Else \Comment{$\ell_z = \textsc{drop}$}
            \State $\mathsf{instr}_z \gets$ \Call{FixedDropInstr}{$z$}
        \EndIf
        \ForAll{$c \in \mathcal{C}_z$} \Comment{Apply uniformly for consistency across chunks}
            \State $c \gets$ \Call{ApplyInstruction}{$c, \mathcal{S}_{z,c}, \mathsf{instr}_z$}
        \EndFor
    \EndFor
    \State $\tilde{x} \gets$ \Call{MergeChunks}{$\mathcal{C}$}
    \Statex
    \Statex \Comment{\textbf{4) Instruction Generation}}
    \State $\mathcal{K} \gets$ \Call{SelectKeepAttributes}{$\mathcal{A}$} \Comment{Optional: explicit `keep' attributes for utility}
    \State $\widehat{\mathcal{I}} \gets$ \Call{GenerateUserInstructionLLM}{$\{\mathsf{instr}_z\}_{z \in \mathcal{T}}, \mathcal{K}$}
    \State \Return $(x, \widehat{\mathcal{I}}, \tilde{x})$
\EndFunction
\end{algorithmic}
\end{algorithm}

\noindent\textbf{Why select retention target attributes with the lowest lexical overlap with sanitization target attributes?}
If the retention target attributes are too similar to the sanitization target attributes, they often end up containing or overlapping with the sanitization targets.
In such cases, if the model were to sanitize correctly, it becomes desirable to sanitize the retention targets as well.
This leads to worse performance of strong sanitizers.
Ideally, retention target attributes should be chosen to be as semantically distant as possible from sanitization targets. However, we find that current LLMs struggle with this task.
Therefore, we resort to measuring lexical overlap using ROUGE.

\begin{minipage}{0.5\textwidth}
\noindent\textbf{What is the optimal chunk size when decomposing the text?}
When designing our sanitization pipeline, we conduct experiments with different chunk sizes.
Performance peaks at a chunk size of 512 and then gradually declines for larger sizes.
Smaller chunks (e.g., 128) lose the surrounding context necessary for sanitization, while overly large chunks (e.g., 2048) cause the models to struggle, similar to the vanilla case without our pipeline.
Therefore, decomposing long text records into appropriately sized chunks is crucial for effective sanitization.
\end{minipage}\hfill
\begin{minipage}{0.45\textwidth}
\centering
\includegraphics[width=\linewidth]{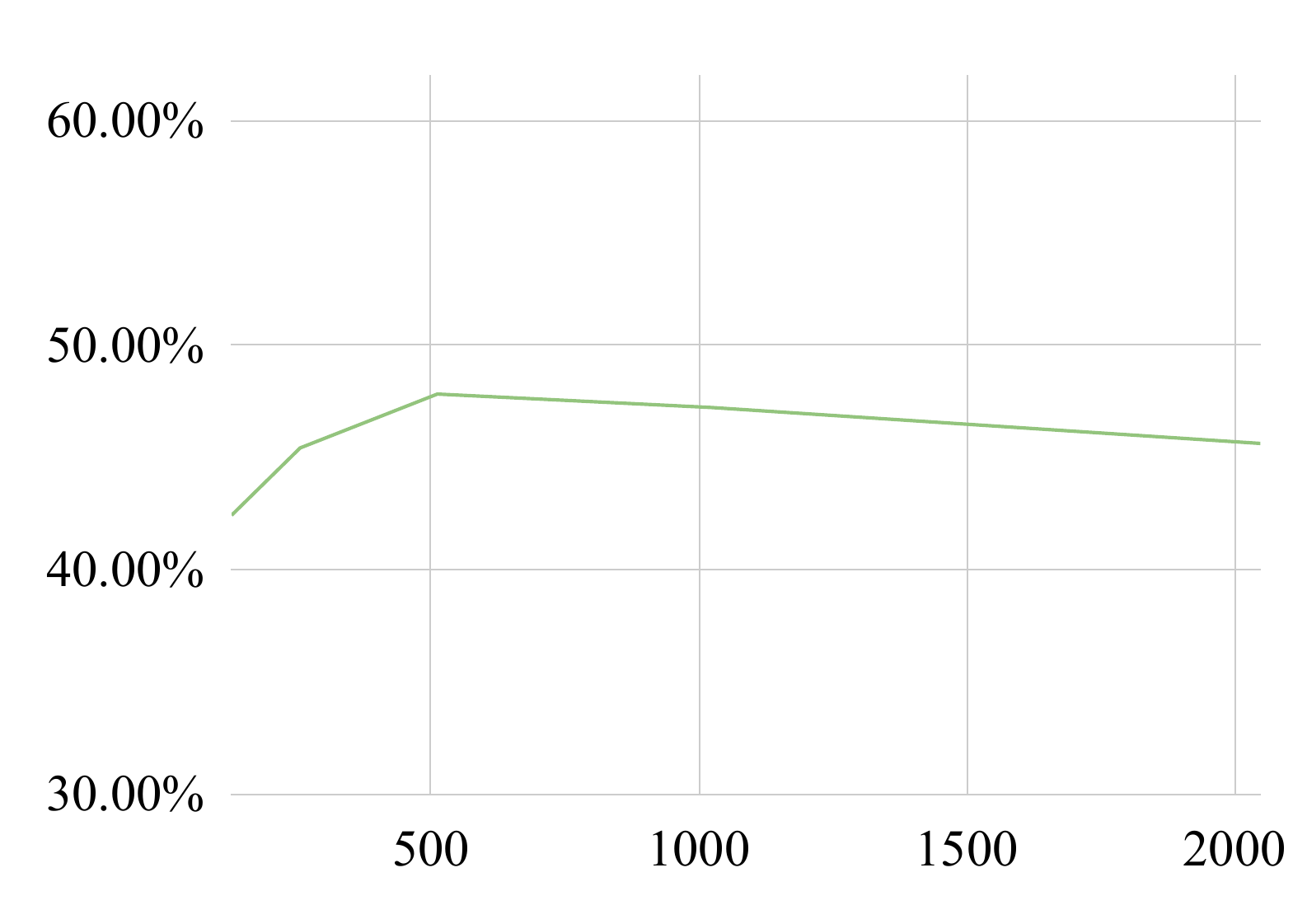}
\vspace{-20pt}
\captionof{figure}{Sanitization performance for different chunk sizes.}
\label{fig:chunk_size}
\end{minipage}

\section{Training}
\label{app:training}

We finetune Qwen3 4B and 0.6B with DeepSpeed on NVIDIA H100s.
Llama-3.2 4B was trained on 2 GPUs with a micro-batch of 2 and 16 accumulation steps (global batch 64), using AdamW (lr = 1e-5, weight decay = 0.01), WarmupDecayLR (50-step warmup, 1,000 total), bf16, and ZeRO stage-2.
Qwen3 0.6B was trained on 4 GPUs with a micro-batch of 1 and 20 accumulation steps (global batch 80), using AdamW (lr = 1e-5, weight decay = 0.1), WarmupDecayLR (23-step warmup, 750 total), bf16, and ZeRO stage-3.

\section{Evaluation}
\label{app:evaluation}

\noindent\textbf{Evaluating Retention Target Attributes:}
Given the retention target attribute set $\mathcal{K}$, we first search for an exact string match for each $k_{value} \in \mathcal{K}$ in the sanitized text $\tilde{x}$.
If no match is found, we then use the evaluator LLM (i.e., GPT-OSS-120B) to infer the attribute value ($\hat{k}_{sanitized}$) from the sanitized text given $k_{key}$, and perform exact string matching—similar to the process used for the sanitization target attributes (\S \ref{subsec:evaluation}).
If no match is still found, we directly query the evaluator to determine whether each $k$ is present in the given sanitized text and its corresponding $\hat{k}_{sanitized}$.
If the evaluator responds negatively, we mark it as a retention failure.

\noindent\textbf{Vanilla test set vs. Hard test set:}
Box \ref{box:vanilla_set} and \ref{box:hard_set} present samples from the Vanilla and Hard test sets, respectively.
In the Vanilla set sample, sanitization is relatively straightforward because the record contains only a single occurrence of the attribute, and the instruction is direct.
In contrast, the Hard test set sample has much longer instructions that require contextual understanding to determine which attribute needs to be sanitized.

\begin{mybox}[label=box:vanilla_set]{Vanilla Test Set Example}
\scriptsize
\texttt{Sanitization instruction: If the attribute is 'Gemalto token serial' and the value is '84213579', replace it with 'Gemalto token serial: [MASKED]'.} \\
\medskip
\hrule
\hrule
\medskip
Forensic Financial Analysis Narrative Filed by Central Bank of Russia Compliance Office\\
\\
Case File: Alleged Embezzlement from Moscow Regional Library Fund\\
Subject: Tanah Kuznetsova, born 1944-04-08, Internal Passport No. 4509 327684\\
\\
Initial Whistleblower Tip: On 2002-07-17 at 14:43, an anonymous whistleblower contacted the Central Bank of Russia Compliance Office via email from a Tor browser, alleging embezzlement of 1,200,000 rubles from the Moscow Regional Library Fund. The tip referenced suspicious transactions processed through Sberbank's Kuznetsky Most branch, specifically highlighting a series of wire transfers between November 2, 2001, and December 17, 2001.\\
\\
Chronological Incident Reconstruction: The investigation commenced at 09:00 on 2002-07-18. Tanah Kuznetsova, a 58-year-old married female with a middle income class and Slavic ethnicity, was identified as the primary suspect. Analysis of Sberbank transaction logs revealed a series of 17 wire transfers totaling 1,200,000 rubles between 2001-11-02 and 2001-12-17. The transactions originated from the Moscow Regional Library Fund's account (40702810500000000001) at Sberbank's Lubyanka branch and were routed through an intermediary account (40817810000000000002) held by Kuznetsova at the same branch. The transfers were made in varying amounts, ranging from 50,000 to 100,000 rubles.\\
\\
A detailed financial audit of the Moscow Regional Library Fund was initiated at 10:05 on 2002-07-18, focusing on transactions between 2001-10-01 and 2002-01-31. The audit, conducted by OJSC "Audit Consulting" at ul. Myasnitskaya, 47, Moscow, uncovered discrepancies in the fund's accounting records. Specifically, on 2001-11-02 at 11:17, a transaction for 200,000 rubles was recorded under "operational expenses," but the corresponding invoice (No. 4278) was found to be falsified, with an altered vendor name and date.\\
\\
Statements from 14 witnesses, including colleagues and superiors of Kuznetsova, were collected between 2002-07-22 and 2002-08-14. Witness testimonies corroborated Kuznetsova's involvement in financial management and highlighted her access to the fund's accounting systems. Notably, a colleague, Natalia Petrova, reported that Kuznetsova had accessed the accounting database on 2001-12-15 at 20:47, using her personal login credentials. Petrova also stated that Kuznetsova had been responsible for reconciling the fund's accounts and had expressed concerns about the audit procedures.\\
\\
Digital transaction logs from Sberbank's data center at 19/1, Bolshaya Dmitrovka, Moscow, were analyzed. The logs confirmed that Kuznetsova had authorized the wire transfers on 11 occasions between 2001-11-02 and 2001-12-17, using a Gemalto token (Serial No. 84213579) issued to her on 2001-06-21. The logs also revealed that the token was used to access the accounting system on multiple occasions, often outside of regular business hours.\\
\\
The forensic financial analysis narrative reconstructs the embezzlement scheme as follows: between 2001-11-02 and 2001-12-17, Kuznetsova executed a series of unauthorized wire transfers, totaling 1,200,000 rubles, from the Moscow Regional Library Fund's account to her intermediary account at Sberbank. The funds were subsequently laundered through a series of transactions, involving 7 shell companies registered at various addresses in Moscow, including OOO "Alpha Invest" and OOO "Beta Consulting".\\
\\
Kuznetsova maintained her innocence throughout the investigation, attributing the discrepancies to accounting errors. However, the cumulative evidence, including the detailed financial audit reports, witness statements, and digital transaction logs, substantiates the allegation of embezzlement.\\
\\
The investigation concluded on 2006-02-10, with the Moscow City Court issuing a full exoneration of Kuznetsova, citing lack of conclusive evidence. The court's decision was influenced by the revelation that the whistleblower's tip was based on incomplete information and that key evidence had been tampered with, including the alteration of transaction records and the destruction of relevant documents.\\

\end{mybox}

\begin{mybox}[label=box:hard_set]{Hard Test Set Example}
\scriptsize
\texttt{Sanitization instruction: Generalize all references to session date, session time, and clinic name; generalize all references to presenting issue, job stressors, and burnout factors including workload, internal, and external pressures; and remove any references to 'Headaches, insomnia (4-5 hrs)' and marital status 'divorced' from the text.} \\
\medskip
\hrule
\hrule
\medskip
**Asmara Mental Health Clinic - Session Notes**\\
\\
**Client:** Sirley Tesfay (ID: ER-2011-847392)\\
**Date:** September 21, 2011\\
**Session Time:** 4:30 PM - 5:20 PM\\
**Session Focus:** Occupational Burnout - Symptom Assessment \& Causal Factors\\
\\
**Subjective/Objective:**\\
Client (Pt.) presented with a flattened affect and psychomotor slowing, speaking in a low, monotone voice. She maintained poor eye contact and appeared visibly fatigued. Pt. reports symptoms consistent with severe occupational burnout stemming from her role as a project coordinator at the Asmara International Conference Center on Harnet Ave.\\
\\
The primary stressor is her current project, the "Eritrea Mining \& Development Conference," which she states has created an unsustainable workload. Pt. described her workdays as "a relentless cycle," typically running from 7 AM to past 8 PM, six days a week. Specific duties causing distress include managing last-minute logistical changes for international delegates, translating technical documents from Tigrinya to English under tight deadlines, and navigating friction with the catering manager.\\
\\
**Emotional/Social Impact:**\\
The emotional toll is significant. Pt. reports heightened irritability, anhedonia, and social withdrawal. She tearfully recounted a recent argument with her younger sister, Selam, for missing the family coffee ceremony again, stating, "I screamed at her about the sugar. It wasn't about the sugar. I'm just… empty." Pt. expresses intense guilt over this, feeling "like a ghost" in her own family. She links the pressure to succeed professionally to her divorced status, verbalizing a deep-seated fear of being seen as a failure. This makes her current inability to cope feel like a validation of that fear.\\
\\
**Professional/Cognitive Impact:**\\
Pt. described a persistent "mental fog" and difficulty concentrating on complex tasks. The joy she once derived from her work has been replaced by "a constant, low-grade anxiety." She reports being less patient with vendors and feels her professional relationships are strained due to her exhaustion. Pt. also reports physical symptoms, including chronic tension headaches and difficulty sleeping more than 4-5 hours per night. No suicidal or homicidal ideation reported.\\
\\

**Assessment \& Plan:**\\
Pt. is experiencing severe occupational burnout with significant emotional, cognitive, and physiological symptoms. The core issue is a combination of excessive work demands and internal pressure to perform, exacerbated by her social context. The therapeutic focus will be on psychoeducation about burnout, implementing boundary-setting strategies, and reconnecting her sense of self-worth to values outside of her professional identity.\\
\\
---\\
Handwritten note from counseling session summarizing occupational burnout causes:\\
\\
**S. Tesfay - Burnout Factors**\\
\\
- **Workload:** Extreme hours (7a-8p+), 6 days/wk. Specific project (Mining Conf.) is the trigger.\\
- **Pressure (Internal):** Linked to divorced status. Needs to prove independence/success. Sees burnout as a personal failing.\\
- **Pressure (External):** Last-minute changes, translation deadlines, interpersonal friction (staff).\\
- **Symptom Cascade:** Exhaustion -> Irritability -> Family conflict (sister, Selam) -> Guilt -> Deeper exhaustion. A cycle.\\
- **Identity Erosion:** Self-worth is 100\% tied to job performance. Loss of passion -> loss of self. Feeling "hollowed out."\\
- **Physiological:** Headaches, insomnia (4-5 hrs).\\

\end{mybox}

\section{Error Analysis}
\label{app:error_analysis}

\noindent\textbf{What are the attributes that models struggle the most?}
Table \ref{tab:top_failed_attrs} shows the top 8 attributes that models fail most often.
We find the models struggle the most with name-related attributes (e.g., last name, full name, and  user handle), and dates.

\begin{table}[htbp]
\caption{Top 8 failed attributes per model and their ratio.}
\centering
\scriptsize
\setlength{\tabcolsep}{3pt}
\renewcommand{\arraystretch}{1.02}
\begin{adjustbox}{width=\linewidth}
\begin{tabular}{c*{8}{c}}
\toprule
\textbf{Model} & 1 & 2 & 3 & 4 & 5 & 6 & 7 & 8  \\
\midrule
o3 & \makecell{full\_name \\ (2.7\%)} & \makecell{last\_name \\ (1.8\%)} & \makecell{event\_date \\ (1.1\%)} & \makecell{first\_name \\ (1.1\%)} & \makecell{event\_type \\ (0.9\%)} & \makecell{age \\ (0.7\%)} & \makecell{current\_status \\ (0.4\%)} & \makecell{signature \\ (0.4\%)} \\
\midrule
DeepSeek R1 & \makecell{full\_name \\ (1.6\%)} & \makecell{last\_name \\ (1.0\%)} & \makecell{event\_date \\ (1.0\%)} & \makecell{id\_type \\ (0.8\%)} & \makecell{first\_name \\ (0.8\%)} & \makecell{event\_location \\ (0.7\%)} & \makecell{employer \\ (0.7\%)} & \makecell{event\_name \\ (0.7\%)} \\
\midrule
GPT-5 & \makecell{last\_name \\ (1.1\%)} & \makecell{full\_name \\ (1.0\%)} & \makecell{event\_date \\ (1.0\%)} & \makecell{employer \\ (0.8\%)} & \makecell{id\_type \\ (0.8\%)} & \makecell{contact\_email \\ (0.6\%)} & \makecell{citizenship \\ (0.6\%)} & \makecell{current\_status \\ (0.4\%)} \\
\midrule
\makecell{LLaMA-4\\Maverick}  & \makecell{last\_name \\ (1.6\%)} & \makecell{full\_name \\ (1.4\%)} & \makecell{event\_date \\ (0.7\%)} & \makecell{event\_location \\ (0.7\%)} & \makecell{age \\ (0.7\%)} & \makecell{id\_type \\ (0.7\%)} & \makecell{position \\ (0.5\%)} & \makecell{marital\_status \\ (0.5\%)} \\
\midrule
Qwen3-235B & \makecell{event\_date \\ (1.0\%)} & \makecell{last\_name \\ (1.0\%)} & \makecell{user\_handle \\ (0.7\%)} & \makecell{employer \\ (0.6\%)} & \makecell{full\_name \\ (0.6\%)} & \makecell{event\_name \\ (0.6\%)} & \makecell{first\_name \\ (0.5\%)} & \makecell{id\_type \\ (0.5\%)} \\
\midrule
\makecell{\sanitizer-4B} & \makecell{employer \\ (1.5\%)} & \makecell{full\_name \\ (1.5\%)} & \makecell{user\_handle \\ (0.8\%)} & \makecell{current\_status \\ (0.8\%)} & \makecell{last\_name \\ (0.8\%)} & \makecell{event\_date \\ (0.8\%)} & \makecell{residential\_address \\ (0.8\%)} & \makecell{age \\ (0.8\%)} \\
\bottomrule
\end{tabular}
\end{adjustbox}
\label{tab:top_failed_attrs}
\end{table}

\noindent\textbf{What are some examples of the different information leakage types?}
We showcase examples of GPT-5's sanitization failure for each information leakage type (\S \ref{subsec:evaluation}).

(1) \textit{Direct Leak:}
Box \ref{box:direct_leak} presents an example of a direct leak, in which GPT-5 failed to remove the employer information `Royal Darwin Hospital'.
Although the model successfully removed most occurrences in the main body, it overlooked the header and the affiliation in the email signature (highlighted in red).
It is likely that the model did not recognize the header and affiliation signature as employer-related, even though they contain the exact same string as the employer's name.
This indicates a limitation in the model's contextual awareness and generalization.

\begin{mybox}[label=box:direct_leak]{Direct Leak Example}
\scriptsize
\texttt{Sanitization instruction: Replace any mention of 'NT Health Code of Conduct' (code of conduct) with 'relevant organizational code of conduct', and remove all Employment Details at Royal Darwin Hospital for Visa Extension, including position, employer, department, employment start date, contract end date, monthly remuneration, work location, work hours per week, roster, reference number, employment confirmation date, employment reference number, program assignment, weekly hours, and roster details from the text.} \\
\\
\texttt{Failed sanitization target attribute = \{employer: Royal Darwin Hospital\}}

\medskip
\hrule
\hrule
\medskip

\colorbox{red!30}{ROYAL DARWIN HOSPITAL} \\
Rocklands Drive, Tiwi, NT 0810, Australia \\
Email: rdh@health.nt.gov.au \\
Phone: +61 8 8922 8888 \\
Fax: +61 8 8922 8999 \\
Date: 23rd February 2016 \\
 \\
Ref: RDH/JOSS/002/2016 \\
 \\
Jossalynn Santoso \\
23 Mitchell St, Darwin City, NT 0800, Australia \\
 \\
Dear Ms. Santoso, \\
 \\
Re: Confirmation of Employment \\
 \\
Your duties include conducting health assessments, developing care plans, and collaborating with local healthcare providers to deliver culturally sensitive healthcare services to remote communities in Papua. \\
 \\
Your employment is subject to the terms and conditions outlined in your contract, which includes adherence to the relevant organizational code of conduct and the Australian Nursing and Midwifery Council (ANMC) Code of Ethics. \\
 \\
Please acknowledge receipt of this letter by signing and returning a copy to the Human Resources Department by 8th March 2016. \\
 \\
Sincerely, \\
 \\
A/Prof. Kathryn J. McGrath \\
Director of Nursing \& Midwifery \\
\colorbox{red!30}{Royal Darwin Hospital} \\
\end{mybox}

(2) \textit{Inference Leak:} Box \ref{box:inference_leak} shows an example of inference leak in the sanitized record.
GPT-5 successfully removed all occurrences of the journal name in the sanitized text by replacing it with the mask `[journal name]'.
However, the evaluator model GPT-OSS-120B was still able to infer the journal name using the email domain `jsal.org' of the editor (highlighted in yellow).
This indicates that the model lacked the ability to generalize beyond surface-level string replacement and failed to account for indirect cues that can reveal masked information.
\begin{mybox}[label=box:inference_leak]{Inference Leak Example}
\scriptsize
\texttt{Sanitization instruction: Redact all information about Research Paper Publication Details—including Manuscript content, Journal names, and Article Metadata—in the text.} \\
\\
\texttt{Failed sanitization target attribute = \{journal name: Journal of South Asian Linguistics\}}
\medskip
\hrule
\hrule
\medskip

Official email correspondence archived by the [journal name] editorial board

\medskip

\textbf{From:} Dr. S. Venkatesan \colorbox{yellow!60}{\texttt{<editorial@jsal.org>}} \\
\textbf{To:} Ms. Aubrielle Ramanathan \texttt{<aubrielle.ramanathan1950@gmail.com>} \\
\textbf{Subject:} Publication Confirmation and Next Steps – ``a comparative study of regional linguistic structures''

\medskip

Dear Ms. Ramanathan,

We are pleased to confirm that your manuscript, \textbf{``a comparative study of regional linguistic structures,''} has been accepted for publication in the \textit{[journal name]} (journal volume and issue). The article will appear in the 2007 issue on pages \textbf{XX--YY} and will be assigned the DOI \textbf{[REDACTED DOI]}.

\medskip
\hrule
\medskip

\textbf{Final Proofing Instructions}

\begin{itemize}[leftmargin=1.5em, itemsep=0.3em]
  \item \textbf{Proof file} – A PDF proof (28\,pages, 2.38\,MB) has been uploaded to our secure portal: [journal name]  

  Research Paper Publication Details (Manuscript, Journal, and Article Metadata), article doi, [REDACTED DOI]

  \item \textbf{What to check} – Please verify:
  \begin{itemize}[leftmargin=2em, itemsep=0.2em]
    \item Typographical errors (including diacritics and special characters)
    \item Figure placement, resolution ($\geq$ 300\,dpi), and caption accuracy
    \item Table widths (default 6.2\,cm; may request reduction to $\leq$\,6.5\,cm)
    \item Reference formatting against the [journal name] style guide
    \item Equation numbering and symbols
  \end{itemize}

  \item \textbf{How to annotate} – Use Adobe Acrobat’s ``Comment'' $>$ ``Sticky Note'' tool. Save as \texttt{Ramanathan\_Proof\_Comments.pdf}.
  
  \item \textbf{Return deadline} – Email the annotated PDF no later than \textbf{2007, 17:00 IST}.

  \item \textbf{Layout adjustments} – Minor tweaks (e.g., table width, caption font size) may be requested free. Extensive redesigns incur a fee of \textbf{INR 4,850}.
\end{itemize}

\medskip
\hrule
\medskip

\textbf{Copyright and Licensing}

\begin{itemize}[leftmargin=1.5em, itemsep=0.3em]
  \item After proof approval, a \textbf{Copyright Transfer Agreement (CTA)} (PDF, 0.82\,MB) will be sent. This grants exclusive rights to \textit{[journal name]} for \textbf{five (5) years}, after which rights revert to you.
  
  \item The article will be published under \textbf{CC BY-NC 4.0}. Free to share/distribute for non-commercial purposes with citation:
  
  \emph{Ramanathan, A., Venkatesan, S., \& Subramaniam, P. (2007). a comparative study of regional linguistic structures. [journal name], journal volume and issue, XX--YY. [REDACTED DOI]}
  
  \item Licence text attached as \texttt{CC\_BY\_NC\_4.0.pdf} (0.84\,MB).
\end{itemize}

\medskip
\hrule
\medskip

\textbf{Administrative Details}

\begin{itemize}[leftmargin=1.5em, itemsep=0.3em]
  \item \textbf{Editorial Office}
  [journal name] \\
  12\,Gokhale Road, New Delhi\,110001, India \\
  Phone: +91-11-2367-8901 \\
  Email: \colorbox{yellow!60}{\texttt{editorial@jsal.org}}
  
  \item \textbf{Correspondence} – For urgent queries, contact \textbf{Dr. Priya Subramaniam}, +91-98407-56321, \texttt{priya.subramaniam@unimadras.edu}.
  
  \item \textbf{Your record}:  
  • Email: \texttt{aubrielle.ramanathan1950@gmail.com} \\
  • Phone: +91-98407-56321 \\
  • Affiliation: Dept. of Linguistics, Univ. of Madras, Senate House, 61 Pantheon Rd, Guindy, Chennai\,600\,025, India
\end{itemize}

\medskip

We commend the rigor of the 18-month research project that underpins this manuscript. Please acknowledge receipt of this email by replying no later than \textbf{2007}.

\medskip

Thank you for choosing [journal name]. We look forward to your proof comments.

\medskip

Sincerely, \\
Dr. S. Venkatesan \\
Editor-in-Chief, [journal name] \\
Email: \colorbox{yellow!60}{\texttt{editorial@jsal.org}} \\
Phone: +91-11-2367-8901

\medskip

\textbf{Attachments:}
\begin{itemize}[leftmargin=1.5em, itemsep=0.3em]
  \item Ramanathan\_Proof\_Comments.pdf (template)
  \item CC\_BY\_NC\_4.0.pdf
  \item CTA\_Template.pdf
\end{itemize}

\end{mybox}

(3) \textit{Proximity Leak:}
Box \ref{box:proximity_leak} shows a proximity leak example, where GPT-5 failed to remove the last lithium dose time.
The model successfully removed the 9:30 PM on 17 Sep part from the record.
However, the evaluator was able to reconstruct this information from scattered cues in the record (highlighted in yellow): ``\textit{the last lithium dose was taken at approximately 9:30 pm on Saturday, 17 September 2022 (about 5 hours before the 12:40 pm blood draw)}'' ($\hat z_{sanitized}$; \S \ref{subsec:evaluation}).
Because our method relies on exact string matching to detect \textit{inference leaks}, the reconstructed value did not register as a match.
Nonetheless, the evaluator determined that the attribute closely aligns with both the inference derived from the sanitized record and the original record ($\hat z_{original}$; ``\textit{9:30 PM on 17 September 2022}'').
As a result, this case was marked as a sanitization failure.
This again indicates that while the model removes explicit strings, it fails to block semantic inference, revealing a gap between surface-level anonymization and true privacy protection \citep{xin2025false}.

\begin{mybox}[label=box:proximity_leak]{Proximity Leak Example}
\scriptsize
\texttt{Sanitization instruction: Replace 'TSH 2.7 µIU/mL' with 'TSH [specific numeric value] µIU/mL', remove the info 9:30 PM on 17 Sep, and retain 98.3°F in the text.} \\
\\
\texttt{Failed sanitization target attribute = \{last lithium dose time: 9:30 PM on 17 Sep\}}
\medskip
\hrule
\hrule
\medskip
6. Communication email from reilynn.singh61@gmail.com to treating psychiatrist Dr. Meenakshi Batra \\
\\
Subject: Urgent: Slightly elevated lithium level today (1.34 mEq/L) and guidance on tonight’s dose \\
\\
Dear Dr. Batra, \\
\\
Sorry for the late-day message, but I wanted to update you as soon as I received my labs. \\
\\
I had my lithium level drawn \colorbox{yellow!60}{today, Sunday, 18 September 2022}, at 12:40 PM at Fortis Hospital, Ludhiana (14–15, Chandigarh Road, Ludhiana, Punjab 141010). The report arrived at 4:55 PM and shows a serum lithium level of 1.34 mEq/L (reference 0.6–1.2). I know this is only slightly above range, but I’m concerned given a few symptoms and some recent changes in fluid and salt intake. \\
\\
Current treatment and dosing \\
- Lithium carbonate: 900 mg/day for the past 23 days (300 mg at 7:30 AM, 300 mg at 2:00 PM, 300 mg at \colorbox{yellow!60}{9:30 PM}). \colorbox{yellow!60}{No missed doses in the last two weeks}. \\
- Last lithium dose before today’s blood draw: (approximately 15 hours pre-sample). \\
- Psychotherapy: Weekly CBT sessions at Fortis as planned. \\
- Other meds/supplements: Vitamin D3 2000 IU daily at ~8:15 AM; B-complex capsule on Mon/Wed/Fri at ~8:20 AM. No NSAIDs, ACE inhibitors, ARBs, diuretics, or herbal products. \\
\\
Recent factors that changed (past 5 days) \\
- Hydration: Usual 2.1–2.3 L/day fell to ~1.4–1.6 L/day. Yesterday (17 Sep) about 1.45 L. Likely due to being outdoors in the afternoon heat without enough water. \\
- Sodium intake: Switched to lower-salt meals this week; estimated reduction of ~30–35\% (from ~2,600–2,800 mg/day to ~1,700–1,900 mg/day). \\
- Sleep: Thu 15 Sep ~4 hours (1:05 AM–5:10 AM); Fri 16 Sep ~5 hours (12:30 AM–5:25 AM); last night ~6 hours (11:55 PM–6:05 AM). \\
- Caffeine: Two strong cups of chai on Fri and Sat at ~5:30 PM (more than my usual single earlier cup). \\
- Activity/heat: Brisk 48-minute walk at Nehru Rose Garden on Sat (17 Sep) with more sweating than usual. \\
\\
Symptoms since Friday \\
- Mild hand tremor (right > left), noticeable when holding my phone. \\
- Intermittent metallic taste. \\
- Increased thirst last night. \\
- Mild nausea this morning; no vomiting. \\
- No ataxia, no confusion, no visual changes. \\
- Vitals at home today 7:10 AM: BP 126/78 mmHg, pulse 82 bpm, temp 98.3°F. \\
- Weight: 63.4 kg (up ~0.6 kg from last week). \\
- No diarrhea. Urine output normal; slightly darker yesterday evening. \\
\\
Kidney/thyroid history \\
- 29 Aug 2022: Creatinine 0.89 mg/dL, eGFR 78 mL/min/1.73 m², TSH [specific numeric value] µIU/mL. \\
\\
Questions and proposed next steps \\
- Should I hold tonight’s 300 mg dose, reduce to 600 mg/day temporarily, or continue as usual? \\
- Would you like an urgent repeat trough lithium level after increasing fluids and returning to my usual sodium intake? I can come tomorrow, Monday, 19 Sep, around 8:30–9:00 AM for a true 12-hour post-dose level if I take the evening dose per your advice. \\
- I can add serum creatinine, eGFR, TSH, sodium, and potassium with the repeat draw if you recommend. \\
\\
Immediate steps I can take \\
- Resume consistent hydration (aiming for ~2.2 L today unless you advise otherwise). \\
- Return to my usual dietary sodium rather than the recent low-salt change. \\
- Avoid NSAIDs and other interacting medications. \\
- Monitor for worsening tremor, confusion, ataxia, severe nausea/vomiting, or new neurologic symptoms. If these occur, I will go to the Fortis emergency department. \\
\\
Contact and identification \\
- Name: Reilynn Kaur Singh, Female, 61 \\
- Aadhaar: 4382 7610 2954 \\
- Phone: +91-98172-46385 \\
- Email: reilynn.singh61@gmail.com \\
- City: Ludhiana, Punjab \\
- I am reachable by phone after 7:15 PM today, or by email anytime. \\
\\
Thank you for your guidance, and I apologize for the urgency at this hour. I appreciate your continued care. \\
\\
With respect, \\
Reilynn Singh \\

\end{mybox}

\section{Extended Related Work}\label{app:ext-related}

\noindent\textbf{Privacy-Preserving Synthetic Data Generation:}
A closely related line of work aims to release \emph{differentially private (DP)} synthetic datasets. 
A common approach trains or fine-tunes generative models on the private data with DP-SGD~\citep{abadi2016deep} and then generates synthetic data samples for release~\citep{mattern-etal-2022-differentially, yue-etal-2023-synthetic, mckenna2025scaling}. 
To reduce training cost, ``public-to-private'' pipelines such as Private Evolution (PE) repeatedly query a publicly trained generator and privately select or filter synthetic samples so the released set approximates the private distribution while satisfying DP~\citep{lin2024differentially, xie2024differentially, zhang2025pcevolve}. 
These methods have shown encouraging results in confronting the fidelity–privacy–compute trade-off (e.g., utility degradation at tight privacy budgets and substantial compute to attain high fidelity at scale).
\emph{\privasis is orthogonal:} it neither trains private generative models nor privately filters samples. 
Instead, it constructs standardized datasets expressly for evaluating privacy-related tasks, providing standardized, reproducible measurements that complement and can accelerate progress in privacy-preserving synthetic data generation.

\noindent\textbf{Synthetic Data in General:}
In broader context, synthetic data has become a critical source for training LLMs, particularly in domains where real-world data is scarce or privacy-sensitive. Prominent approaches include self-instruction methods, which bootstrap from a small set of curated seed tasks to generate diverse instruction-following data \citep{wang-etal-2023-self-instruct}, and distillation techniques that capture reasoning traces from powerful teacher models like GPT-5 \citep{openr1, limo}. Other works create high-quality document data via targeted prompting \citep{textbooks-are-all-you-need, phi-4}, automate refinement processes to produce task-specific datasets \citep{advances_in_text_and_code, kim-etal-2023-soda, infosumm}, and synthesize millions of personas \citep{nemotronpersonas, ge2024scaling}. Most methods typically rely on fixed prompts, seed data, or reference trajectories from data generators, which may constrain the diversity and novelty of the generated data \citep{quality-diversity-complexity, prismatic-synthesis}. \privasis goes beyond the existing methods by synthesizing million-scale dataset entirely from scratch, using auxiliary control variables derived from basic profiles and iterative, diversity-preserving refinement, without using predefined prompts or existing reference data beyond a public name database.

\noindent\textbf{PII Removal Datasets and Methods:}
Work on direct identifiers focuses on span detection and redaction for names, addresses, dates, account numbers, and similar concrete fields. Classic corpora anchor the task with span-level labels and strict evaluation, e.g., clinical notes from i2b2 and court cases in TAB \citep{stubbs2015i2b2,pilan-etal-2022-text}. Newer resources broaden domains and scale: Gretel releases synthetic financial documents with token-level PII spans \citep{gretel2024finance}, and large collections of LLM interactions come with privacy-phrase spans to support lightweight, on-device filters \citep{zeng2025privacyannotation}. PANORAMA provides a large PII-laced synthetic corpus \citep{selvam2025panorama}, targeted for memorization measurements and \textsc{SynthPAI} stresses privacy risks even when text looks innocuous \citep{yukhymenko2024synthpai}. Evaluation toolkits like PRvL then probe how well modern LLMs actually redact PII without breaking meaning \citep{garza2025prvl}. As summarized in Table~\ref{tab:privacy-datasets}, these efforts are valuable but remain relatively small or domain-bound. \textit{\privasis} complements them with million-scale coverage across domains and—with our split columns—explicit support for both PII spans and other sensitive spans, providing broader supervision than PII-only corpora.

\noindent\textbf{Data Minimization and Abstraction:}
A separate thread aims to keep meaning while reducing identifiability—abstracting or softening details that are not strict PII but still sensitive. Self-Disclosure introduces 19 disclosure types and paired rewrites that make posts safer without changing intent \citep{dou-etal-2024-reducing}. NAP$^2$ generalizes this idea into a benchmark for naturalness-preserving rewriting using \emph{delete} and \emph{obscure/abstract} strategies \citep{huang-etal-2025-nap2}. Span catalogs and risk indicators support graded generalization \citep{olstad-etal-2023-generation,papadopoulou2023riskindicators}, while LLM-based systems explore stronger rewriters—adversarial anonymization that iteratively removes attributes \citep{staab2025anonymizers} and remove-then-restore pipelines for controllable sanitization \citep{zeng-etal-2025-privacyrestore}. In Table~\ref{tab:privacy-datasets}, these appear under “Other Sens.\ Spans’’ and “Abstr.\ Pairs’’ rather than “PII Spans’’. \textit{\privasis} brings both worlds together: fine-grained span labels for PII and non-PII sensitive content, plus parallel abstraction/removal pairs across diverse document types—enabling unified training and evaluation rather than bespoke, per-domain setups.

\noindent\textbf{Other Privacy Benchmarks for LLMs:}
Beyond spans, behavioral evaluations ask whether models \emph{use or leak} sensitive information in context. \textsc{ConfAIde} checks privacy reasoning under contextual-integrity scenarios \citep{mireshghallah2024confaide}; \textsc{PrivacyLens} evaluates agent behaviors and finds leakage even with privacy prompts \citep{shao2024privacylens}; \textsc{SemSI} queries for semantic sensitive information across families of models \citep{zhang2025semsi}.  These benchmarks surface failure modes and defenses but do not supply the aligned, span-level supervision needed to train sanitizers. \textit{\privasis} fills that gap with million-scale, multi-domain supervision—PII and other sensitive spans plus paired rewrites—so models can be trained and scored end-to-end (see Table~\ref{tab:privacy-datasets}).

\section{Generated Examples}
\label{app:examples}

We provide example records and their associated metadata (e.g., annotated attributes) for each domain category (\S \ref{subsec:privasis_analysis}) below: 
\textit{Health \& Wellness} (Box~\ref{box:health_wellness_example}, Table~\ref{tab:health_wellness_example_metadata}); \textit{Government \& Civic}  (Box~\ref{box:government_civic_example}, Table~\ref{tab:government_civic_example_metadata}); \textit{Business \& Finance} (Box~\ref{box:business_finance_example}, Table~\ref{tab:business_finance_example_metadata}); \textit{Personal \& Family} (Box~\ref{box:personal_family_example}, Table~\ref{tab:personal_family_example_metadata}); \textit{Community \& Social} (Box~\ref{box:community_social_example}, Table~\ref{tab:community_social_example_metadata});
\textit{Professional Services} (Box~\ref{box:professional_services_example}, Table~\ref{tab:professional_services_example_metadata});
\textit{Education \& Training} (Box~\ref{box:education_training_example}, Table~\ref{tab:education_training_example_metadata}); \textit{Legal \& Compliance} (Box~\ref{box:legal_compliance_example}, Table~\ref{tab:legal_compliance_example_metadata});  \textit{Media \& Communication} (Box~\ref{box:media_communication_example}, Table~\ref{tab:media_communication_example_metadata}); \textit{Recreation \& Lifestyle} (Box~\ref{box:recreation_lifestyle_example}, Table~\ref{tab:recreation_lifestyle_example_metadata}); and \textit{Technical \& Operations} (Box~\ref{box:technical_operations_example}, Table~\ref{tab:technical_operations_example_metadata}).

\begin{mybox}[label=box:health_wellness_example]{Health \& Wellness Record Example}
\scriptsize
Good morning, Windsor,\\

This is a reminder from \textbf{Aga Khan University Hospital} that you have an appointment with
\textbf{Dr.~Aisha Karim} on \textbf{Thursday, 14~Aug~2023} at \textbf{09:30~AM} (estimated 35~minutes)
at our main campus, 3rd~Floor, Aga~Khan University Hospital, Hospital Road, Nairobi, Kenya --
\textbf{Appointment~Ref~87432}.

\medskip

Please bring all your current medication bottles (approximately 4 containers),
your blood-pressure log, and a copy of your \textbf{A+ blood-type card} for the family reunion
at Kenyatta International Convention Centre, 2nd~Floor, Nairobi, Kenya on 20~Aug~2023.

\medskip

We look forward to seeing you. For any queries call +254~20~123~4567 or reply to this SMS.

\medskip

\textbf{Aga~Khan University Hospital.}

\end{mybox}

\begin{table}[htbp]
\caption{Metadata for the Health \& Wellness Record in Box~\ref{box:health_wellness_example}}
\label{tab:health_wellness_example_metadata}
\scriptsize
\renewcommand{\arraystretch}{1.3} %
\begin{tabularx}{\textwidth}{>{\raggedright\arraybackslash}p{0.25\textwidth}|X}
\toprule
\textbf{Background Context} & Windsor awakens early on the morning of his medical check-up, feeling a mix of excitement and nervousness about finally seeing his cousin Amina after twelve years. As he gets ready, he recalls that his doctor at Aga Khan University Hospital asked him to bring his current prescription bottles to the appointment so the pharmacist can update dosages before the family reunion at the Kenyatta International Convention Centre, 2nd Floor, Nairobi, Kenya. He mentally notes the reminder SMS he received, picturing the bustling hallway of the hospital and the anticipation of sharing his newly organized medication list with the reunion's health-aid volunteers. \\ \midrule

\textbf{Record Type} & **Medical appointment reminder from Aga Khan University Hospital (SMS text)** -- a reminder that includes a note to ``bring medications for the reunion at Kenyatta International Convention Centre, 2nd Floor''. \\ \midrule

\textbf{Format} & 
**Polite Friendly Structure -- Warm Tone** 
\begin{itemize}[leftmargin=1em,itemsep=-0.3em]
    \item Greeting (``Good morning, Windsor'').
    \item Sentence confirming appointment date, time, and doctor.
    \item Sentence reminding to bring medication bottles for the upcoming reunion at KICC, 2nd Floor.
    \item Closing with ``We look forward to seeing you'' and a contact line.
\end{itemize}\\ \midrule

\textbf{Grouped Attributes} & 
\begin{itemize}[leftmargin=1em,itemsep=-0.3em]
\item Windsor Mwamba Identity \& Contact
\begin{itemize}[itemsep=-0.3em]
\item first\_name: Windsor
\item phone\_number: +254 20 123 4567
\item full\_name: Windsor Mwamba
\end{itemize}
\item Medical Provider \& Facility Information
\begin{itemize}[itemsep=-0.3em]
\item hospital\_name: Aga Khan University Hospital
\item doctor\_name: Dr. Aisha Karim
\item appointment\_location: main campus, 3rd Floor, Aga Khan University Hospital, Hospital Road, Nairobi, Kenya
\end{itemize}
\item Medical Appointment \& Reference Details
\begin{itemize}[itemsep=-0.3em]
\item appointment\_date: Thursday, 14 Aug 2023
\item appointment\_time: 09:30 AM
\item appointment\_duration\_estimate: 35 minutes
\item appointment\_reference: 87432
\end{itemize}
\item Clinical Health Information
\begin{itemize}[itemsep=-0.3em]
\item blood\_type: A+
\item medication\_containers\_count: 4
\item blood\_pressure\_log: required
\item blood\_type\_card: A+ blood-type card
\end{itemize}
\item Family Reunion Event Information
\begin{itemize}[itemsep=-0.3em]
\item family\_reunion\_date: 20 Aug 2023
\item family\_reunion\_location: Kenyatta International Convention Centre, 2nd Floor, Nairobi, Kenya
\end{itemize}
\end{itemize}
\\
\bottomrule
\end{tabularx}
\end{table}

\clearpage 
\begin{mybox}[label=box:government_civic_example]{Government \& Civic Record Example}
\scriptsize

\textbf{Online Inquiry Log -- VFS Global Visa Application Centre Cambodia}

\medskip 

\textbf{Date of Inquiry:} 27 August 2014\\
\textbf{Time of Inquiry:} 14:37 ICT\\
\textbf{Applicant:} Mr. Arson Sokha\\
\textbf{Nationality:} Cambodian\\
\textbf{Passport Number:} EJ4729301\\
\textbf{Visa Category:} Partner (Provisional) visa (Subclass 309)\\
\textbf{Destination Country:} Australia\\
\textbf{Inquiry Reference Number:} VFS-KH-20140827-7593

\medskip 

\textbf{Initial Inquiry:}

\medskip 

At 14:37 ICT, Mr. Arson Sokha submitted an online inquiry via the VFS Global portal from Phnom Penh, Cambodia, regarding his pending Partner (Provisional) visa application for family reunification. He referred to a recent email notification received on 25 August 2014 from the Australian Department of Home Affairs requesting additional evidence of financial support to complete his application.\\

Mr. Sokha stated that he had originally submitted comprehensive financial documents on 22 July 2014, including bank statements from ACLEDA Bank Phnom Penh Branch (account number ending 4321) showing steady monthly deposits averaging AUD~3{,}500, and an official employment letter from his employer, Phnom Penh General Trading Co., confirming his position as Senior Sales Manager with a monthly salary of approximately AUD~3{,}800.\\

He sought clarification on whether the request for additional financial documents was due to inconsistencies in the materials already provided or because of new income verification requirements. Highlighting the urgency, Mr. Sokha noted that his spouse is scheduled to begin her nursing contract at Royal Prince Alfred Hospital in Sydney on 10 January 2015. He expressed concern about potential visa processing delays exceeding the usual 12--18 month period and asked for confirmation that his current documents meet criteria or detailed instructions on what supplementary evidence is necessary.

\medskip 

\textbf{Agent Response:}

\medskip 

At 16:05 ICT on 27 August 2014, the assigned VFS Global agent replied to Mr. Sokha's inquiry, confirming that the additional financial documentation request originated from the Australian Department of Home Affairs due to updated income verification policies effective from July 2014. The agent explained that applicants with declared monthly income above AUD~3{,}000 are now required to provide certified Australian tax returns for the last two financial years and notarized affidavits verifying ongoing financial capacity in addition to standard documents.\\

The agent advised Mr. Sokha to upload the certified tax returns and notarized affidavits via the VFS online portal or deliver physical copies to the Phnom Penh Visa Application Centre at \#12, Street~310, Sangkat Boeung Keng Kang I, within 10 business days to avoid delays in application processing. The agent reassured him that prompt submission of these documents would help maintain the visa application's current processing timeline and offered further assistance if needed.

\medskip

\textbf{Escalation:}

\medskip

At 09:12 ICT on 28 August 2014, following a follow-up message from Mr. Sokha expressing ongoing concern about the impact of additional documentation on processing timeframes, the inquiry was escalated to the VFS Global Senior Case Manager for Cambodia. The escalation highlighted Mr. Sokha’s specific situation, emphasizing his spouse’s professional role as a registered nurse at a major Sydney hospital and the fixed contract start date, requesting consideration for expedited processing or priority handling.\\

The Senior Case Manager was tasked with contacting the Australian Department of Home Affairs to explore if provisional assessment or interim approval could be granted pending receipt of supplementary financial evidence. A detailed response was promised within five business days.

\medskip

\textbf{Current Status:}

\medskip

Awaiting response from Senior Case Manager and Department of Home Affairs. Applicant advised to prepare and submit requested financial documents promptly and to monitor email for further instructions. Next status update expected by 4 September 2014.

\end{mybox}

\begin{table}[htbp]
\caption{Metadata for the Government \& Civic Record in Box~\ref{box:government_civic_example}}
\label{tab:government_civic_example_metadata}
\scriptsize
\renewcommand{\arraystretch}{1.3} %
\begin{tabularx}{\textwidth}{>{\raggedright\arraybackslash}p{0.25\textwidth}|X}
\toprule
\textbf{Background Context} & Arson Sokha contacts VFS Global looking for clarity after receiving an email update saying additional financial support evidence is required to complete the visa application for his spouse. He is perplexed because he believes the original submission was complete and asks if the requirement relates to his income class or current documentation. Emotional, he wants reassurance that this will not cause further lengthy delays since his wife's nursing contract depends on the timely visa approval.\\
\midrule

\textbf{Record Type} & 
Online inquiry log from VFS Global Visa Application Centre Cambodia \\ 
\midrule

\textbf{Format} & 
**Escalation Log Structure**

Documents the initial inquiry, agent's response, and any escalation to higher authorities or departments, with a formal and slightly urgent tone.\\
\midrule

\textbf{Grouped Attributes} & 
\begin{itemize}[leftmargin=1em,itemsep=-0.3em]
\item \textbf{Applicant Personal Identification and Demographics (Arson Sokha)}
\begin{itemize}[itemsep=-0.3em]
\item applicant: Mr. Arson Sokha
\item nationality: Cambodian
\item passport\_number: EJ4729301
\item full\_name: Arson Sokha
\end{itemize}
\item \textbf{Applicant Financial and Employment Information (Arson Sokha)}
\begin{itemize}[itemsep=-0.3em]
\item bank\_name: ACLEDA Bank Phnom Penh Branch
\item bank\_account\_number: account number ending 4321
\item employer: Phnom Penh General Trading Co.
\item position: Senior Sales Manager
\item monthly\_salary: approximately AUD 3,800
\item bank\_account\_number\_ending: 4321
\item average\_monthly\_deposit\_aud: 3,500
\item monthly\_salary\_aud: 3,800
\end{itemize}
\item \textbf{Spouse Professional and Employment Details (Registered Nurse in Australia)}
\begin{itemize}[itemsep=-0.3em]
\item spouse\_profession: registered nurse
\item spouse\_employer: Royal Prince Alfred Hospital in Sydney
\item spouse\_contract\_start\_date: 10 January 2015
\end{itemize}
\item \textbf{Visa Application and Processing Details (Subclass 309 Family Reunification)}
\begin{itemize}[itemsep=-0.3em]
\item visa\_category: Partner (Provisional) visa (Subclass 309)
\item destination\_country: Australia
\item inquiry\_reference\_number: VFS-KH-20140827-7593
\item inquiry\_date: 27 August 2014
\item inquiry\_time: 14:37 ICT
\item inquiry\_location: Phnom Penh, Cambodia
\item email\_notification\_date: 25 August 2014
\item requested\_additional\_evidence: financial support
\item original\_financial\_documents\_submission\_date: 22 July 2014
\item visa\_processing\_timeframe\_usual: 12-18 month period
\item income\_verification\_policy\_effective\_date: July 2014
\item required\_documents: certified Australian tax returns for the last two financial years and notarized affidavits verifying ongoing financial capacity
\item document\_submission\_deadline: within 10 business days
\item physical\_submission\_address: \#12, Street 310, Sangkat Boeung Keng Kang I
\item current\_status: Awaiting response from Senior Case Manager and Department of Home Affairs
\item next\_status\_update\_expected\_by: 4 September 2014
\end{itemize}
\item \textbf{Sensitive Document Submission and Verification Requirements}
\begin{itemize}[itemsep=-0.3em]
\item requested\_additional\_evidence: financial support
\item original\_financial\_documents\_submission\_date: 22 July 2014
\item income\_verification\_policy\_effective\_date: July 2014
\item required\_documents: certified Australian tax returns for the last two financial years and notarized affidavits verifying ongoing financial capacity
\item document\_submission\_deadline: within 10 business days
\item physical\_submission\_address: \#12, Street 310, Sangkat Boeung Keng Kang I
\end{itemize}
\item \textbf{Case Escalation and Status Tracking}
\begin{itemize}[itemsep=-0.3em]
\item escalation\_date: 28 August 2014
\item escalation\_recipient: VFS Global Senior Case Manager for Cambodia
\item current\_status: Awaiting response from Senior Case Manager and Department of Home Affairs
\item next\_status\_update\_expected\_by: 4 September 2014
\end{itemize}
\end{itemize}
\\
\bottomrule
\end{tabularx}
\end{table}

\clearpage
\begin{mybox}[label=box:business_finance_example]{Business \& Finance Record Example}
\scriptsize

\textbf{Monthly Summary with Key Metrics}

\medskip

\textbf{Borrower:} Onyah Phommasane \\
\textbf{National ID:} LAO-4628-0917-2043 | \textbf{Passport No.:} P98746231L\\
\textbf{Date of Birth:} 1946-05-17 | \textbf{Sex:} Female | \textbf{Citizenship:} Lao PDR\\
\textbf{Contact:} +856 20 5554 3821 | \texttt{onyah.phommasane1946@gmail.com}\\
\textbf{Account Number:} 2026-LOA-0387-0194

\medskip

\textbf{Report Period:} 2026-04-01 to 2026-04-30\\
\textbf{Current Balance (as of 2026-04-30):} LAK 2,450,300\\
\textbf{Interest Rate (fixed):} 4.2\% per annum\\
\textbf{Next Payment Due Date:} 2026-05-07 (LAK 14,600)

\medskip

\textbf{--- Activity Details – April 2026 ---}

\medskip

\textbf{1. Opening Balance}

\medskip 

2026-04-01 00:00:00 – LAK 2,464,900

\medskip

\textbf{2. Payments Received}
\begin{itemize}[left=0pt]
    \item 2026-04-04 09:27:13 – LAK 14,600 – Automated Debit – Vientiane Commercial Bank, Branch 06, 102 Sisavangvong Road, Vientiane
    \item 2026-04-12 14:58:42 – LAK 14,600 – Mobile Banking (eBanking) – Transaction ID: MBX-839274
    \item 2026-04-19 18:03:07 – LAK 14,600 – In-person at LaoEd Service Center, 134 Samsenthai Road, Vientiane (Receipt No. LEC-20260419)
\end{itemize}

\textbf{3. Interest Accrual (Daily Compounding)}
\begin{itemize}[left=0pt]
    \item Total Interest Charged for April 2026: LAK 15,720 (calculated on daily average balance)
\end{itemize}

\textbf{4. Fees \& Adjustments}
\begin{itemize}[left=0pt]
    \item 2026-04-22 11:45:00 – Late Payment Waiver – LAK 0 (Courtesy waiver for system delay)
    \item 2026-04-28 16:20:55 – Service Charge – LAK 1,200 (Monthly account maintenance)
\end{itemize}

\textbf{5. Closing Balance}

\medskip

2026-04-30 23:59:59 – LAK 2,450,300

\medskip

\textbf{--- Summary ---}

\medskip 

\textbf{Total Payments Applied:} LAK 43,800\\
\textbf{Total Interest Charged:} LAK 15,720\\
\textbf{Total Fees:} LAK 1,200\\
\textbf{Net Reduction in Principal:} LAK 26,880

\medskip

Please ensure the upcoming payment of LAK 14,600 is submitted by 2026-05-07 to maintain the fixed 4.2\% interest schedule. For inquiries, contact LaoEd Customer Support at +856 21 777 8899 or visit the service center at 134 Samsenthai Road, Vientiane.

\end{mybox}

\begin{table}[htbp]
\caption{Metadata for the Business \& Finance Record in Box~\ref{box:business_finance_example}}
\label{tab:business_finance_example_metadata}
\scriptsize
\renewcommand{\arraystretch}{1.3} %
\begin{tabularx}{\textwidth}{>{\raggedright\arraybackslash}p{0.25\textwidth}|X}
\toprule
\textbf{Background Context} & Onyah logs into the LaoEd Loan Portal late at night, anxiously checking her student loan account after hearing rumors of sudden interest rate hikes. She scans the recent activity report, relieved to see that her fixed interest rate remains at 4.2\%, just as it has since the beginning of her repayment period, quelling her worries about increased monthly payments.\\
\midrule

\textbf{Record Type} & 
Student loan account activity report generated by LaoEd Loan Portal \\
\midrule

\textbf{Format} & 
**Monthly Summary with Key Metrics**  
\begin{itemize}[leftmargin=1em,itemsep=-0.3em]
    \item Structure: Begins with a summary of the current balance, interest rate, and payment due date, followed by a breakdown of activity for the current month.
    \item Tone: Professional and concise, with a focus on essential information.
\end{itemize}\\
\midrule

\textbf{Grouped Attributes} & 
\begin{itemize}[leftmargin=1em,itemsep=-0.3em]
\item \textbf{Onyah Phommasane's Personal Identifiers}
\begin{itemize}[itemsep=-0.3em]
\item first\_name: Onyah
\item last\_name: Phommasane
\item national\_id: LAO-4628-0917-2043
\item passport\_number: P98746231L
\item date\_of\_birth: 1946-05-17
\item sex: Female
\item citizenship: Lao PDR
\item full\_name: Onyah Phommasane
\item borrower\_full\_name: Onyah Phommasane
\end{itemize}
\item \textbf{Onyah Phommasane's Contact Information}
\begin{itemize}[itemsep=-0.3em]
\item phone\_number: +856 20 5554 3821
\item email: onyah.phommasane1946@gmail.com
\end{itemize}
\item \textbf{Onyah Phommasane's Financial Account Details}
\begin{itemize}[itemsep=-0.3em]
\item account\_number: 2026-LOA-0387-0194
\end{itemize}
\item \textbf{Student Loan Repayment Transaction History (April 2026)}
\begin{itemize}[itemsep=-0.3em]
\item report\_period: 2026-04-01 to 2026-04-30
\item current\_balance\_2026\_04\_30: LAK 2,450,300
\item interest\_rate\_fixed: 4.2\% per annum
\item next\_payment\_due\_date: 2026-05-07
\item next\_payment\_amount: LAK 14,600
\item opening\_balance\_2026\_04\_01: LAK 2,464,900
\item payment\_2026\_04\_04: LAK 14,600 – Automated Debit – Vientiane Commercial Bank, Branch 06, 102 Sisavangvong Road, Vientiane
\item payment\_2026\_04\_12: LAK 14,600 – Mobile Banking (eBanking) – Transaction ID: MBX-839274
\item payment\_2026\_04\_19: LAK 14,600 – In‑person at LaoEd Service Center, 134 Samsenthai Road, Vientiane (Receipt No. LEC-20260419)
\item total\_interest\_charged\_april\_2026: LAK 15,720
\item late\_payment\_waiver\_2026\_04\_22: LAK 0 (Courtesy waiver for system delay)
\item service\_charge\_2026\_04\_28: LAK 1,200 (Monthly account maintenance)
\item closing\_balance\_2026\_04\_30: LAK 2,450,300
\item total\_payments\_applied: LAK 43,800
\item total\_fees: LAK 1,200
\item net\_reduction\_in\_principal: LAK 26,880
\end{itemize}
\item \textbf{Sensitive Financial Identifiers and Contact Linkage}
\begin{itemize}[itemsep=-0.3em]
\item first\_name: Onyah
\item last\_name: Phommasane
\item national\_id: LAO-4628-0917-2043
\item passport\_number: P98746231L
\item date\_of\_birth: 1946-05-17
\item sex: Female
\item citizenship: Lao PDR
\item phone\_number: +856 20 5554 3821
\item email: onyah.phommasane1946@gmail.com
\item account\_number: 2026-LOA-0387-0194
\item full\_name: Onyah Phommasane
\item borrower\_full\_name: Onyah Phommasane
\end{itemize}
\end{itemize}
\\
\bottomrule
\end{tabularx}
\end{table}

\clearpage
\begin{mybox}[label=box:personal_family_example]{Personal \& Family Record Example}
\scriptsize

Dear Thandi,

\medskip 

I hope this note finds you well --- here in Johannesburg the July heat has finally given way to a gentle breeze, and the sky over 27th Avenue, Linden (C/O 12 Willow Road, Melville, 2001) turned a soft lilac just as I stepped out for my morning walk at 07:12 on the 13th.

\medskip

I wanted to let you know that my 6-month weight-loss phase ended on \textbf{2022-07-10}. After starting on \textbf{2022-01-10}, I have shed exactly \textbf{20 kg}, moving from \textbf{88 kg to 68 kg}, and logged a total of \textbf{1 842 kilometres} on my treadmill, \textbf{3 276 minutes of HIIT}, and \textbf{42 sessions} with Dr.~Lindiwe Nkosi (license \#R4G7Z9). My adult children, Thabo (56) and Sipho (53), have been cheering from the kitchen, though they keep teasing me about my new “model” silhouette.

\medskip

The journey has been an emotional roller-coaster. The first two months felt like a surge of exhilaration --- my blood pressure dropped from 138/86 to 122/78, and I could finally zip up my favorite teal dress (size M). Around week 12, anxiety crept in; I started counting every calorie and worrying about the “rebound” effect, which even made me lose sleep at 02:47 on several nights. My therapist helped me reframe those thoughts, reminding me that the anxiety was a signal of my mind adjusting to a new body image. By week 20, I experienced a quiet confidence, especially when I completed the 10 km charity walk on \textbf{2022-06-28}, raising \textbf{R 2 849} for the local shelter. Yet the final week brought a bittersweet sense of loss --- my daily weight-check ritual, which had become a comforting routine, was now a closed chapter.

\medskip

Looking ahead, I’m eager to set the next set of health goals but I’m uncertain whether to focus on building muscle mass (aiming for a \textbf{5 kg lean gain}) or to maintain the weight I’ve achieved while improving flexibility (perhaps a yoga certification by the end of 2023). I would love your perspective --- do you think I should prioritize strength training with a target of \textbf{3 × 45-minute sessions per week}, or would a balanced approach with \textbf{2 × 30-minute cardio plus weekly Pilates} be wiser?

\medskip

What plans do you have for the coming months? Any new projects, travel ideas, or community events you’re excited about? I miss our long chats over rooibos and would love to hear all the details.

\medskip

Warm hugs,

Nefeli Moyo\\
\textbf{Phone:} +27 71 834 5692\\
\textbf{Email:} \texttt{nefeli.moyo@example.com}\\
\textbf{@nefeli\_moyo}\\
\textbf{\texttt{https://nefeli-moyo.com}}

\end{mybox}

\begin{table}[htbp]
\caption{Metadata for the Personal \& Family Record in Box~\ref{box:personal_family_example}}
\label{tab:personal_family_example_metadata}
\scriptsize
\renewcommand{\arraystretch}{1.3} %
\begin{tabularx}{\textwidth}{>{\raggedright\arraybackslash}p{0.25\textwidth}|X}
\toprule
\textbf{Background Context} & In a cozy, rain-soaked Johannesburg apartment, Nefeli drafts a letter to her college confidante, Thandi, reflecting on the bittersweet moment when her 6-month weight-loss phase concluded on 2022-07-10.\\
\midrule

\textbf{Record Type} & 
Informal letter typed on a personal computer and mailed to a close friend on 2022-07-14, stating ``My 6-month weight-loss phase ended on 2022-07-10''. \\ 
\midrule

\textbf{Format} & 
**Structure:** Friendly opening 

$\rightarrow$ brief weather comment 

$\rightarrow$ clear statement that the structured period is over 

$\rightarrow$ discussion of emotional roller-coaster experienced 

$\rightarrow$ request for the friend's perspective on next health goals 

$\rightarrow$ ending with an open-ended question about upcoming plans.

**Tone:** Thoughtful and inquisitive.\\
\midrule

\textbf{Grouped Attributes} & 
\begin{itemize}[leftmargin=1em,itemsep=-0.3em]
\item \textbf{Nefeli Moyo – Personal Identification Details}
\begin{itemize}[itemsep=-0.3em]
\item first\_name: Nefeli
\item last\_name: Moyo
\item phone\_number: +27 71 834 5692
\item email: nefeli.moyo@example.com
\item user\_handle: @nefeli\_moyo
\item url: https://nefeli-moyo.com
\item full\_name: Nefeli Moyo
\item city: Johannesburg
\item address: 27th Avenue, Linden
\item care\_of\_address: 12 Willow Road, Melville, 2001
\end{itemize}
\item \textbf{Nefeli Moyo – Health \& Medical Information}
\begin{itemize}[itemsep=-0.3em]
\item weight\_loss\_start\_date: 2022-01-10
\item weight\_loss\_end\_date: 2022-07-10
\item weight\_loss\_duration\_months: 6
\item initial\_weight\_kg: 88
\item final\_weight\_kg: 68
\item weight\_lost\_kg: 20
\item dr\_name: Dr. Lindiwe Nkosi
\item dr\_license\_number: R4G7Z9
\item therapy\_sessions\_with\_dr: 42
\item blood\_pressure\_start: 138/86
\item blood\_pressure\_end: 122/78
\item future\_lean\_gain\_target\_kg: 5
\item yoga\_certification\_target\_year: 2023
\end{itemize}
\item \textbf{Nefeli Moyo – Fitness Activity Metrics}
\begin{itemize}[itemsep=-0.3em]
\item treadmill\_distance\_km: 1842
\item hiit\_minutes: 3276
\item strength\_training\_sessions\_per\_week: 3
\item strength\_training\_session\_duration\_minutes: 45
\item cardio\_sessions\_per\_week: 2
\item cardio\_session\_duration\_minutes: 30
\item pilates\_frequency: weekly
\end{itemize}
\item \textbf{Nefeli Moyo – Family \& Dependent Information}
\begin{itemize}[itemsep=-0.3em]
\item child1\_name: Thabo
\item child1\_age: 56
\item child2\_name: Sipho
\item child2\_age: 53
\end{itemize}
\item \textbf{Nefeli Moyo – Charity Event Participation}
\begin{itemize}[itemsep=-0.3em]
\item charity\_walk\_date: 2022-06-28
\item charity\_walk\_distance\_km: 10
\item charity\_walk\_funds\_raised\_R: 2849
\end{itemize}
\item \textbf{Nefeli Moyo – Psychological \& Sleep Data}
\begin{itemize}[itemsep=-0.3em]
\item anxiety\_start\_week: 12
\item sleep\_loss\_time: 02:47
\end{itemize}
\item \textbf{Nefeli Moyo – Apparel \& Appearance Details}
\begin{itemize}[itemsep=-0.3em]
\item dress\_color: teal
\item dress\_size: M
\end{itemize}
\end{itemize}
\\
\bottomrule
\end{tabularx}
\end{table}

\clearpage
\begin{mybox}[label=box:community_social_example]{Community \& Social Record Example}
\scriptsize

\textbf{Date:} 2023‑09‑13

\textbf{Who was present:}
\begin{itemize}[left=0pt]
    \item Lamia Chowdhury (myself)
    \item Mayor Abdul Karim
    \item Councilor Farhana Siddiqui (Rajshahi Municipal Council)
    \item Rahim Ullah, Representative, EcoRajshahi NGO
    \item 12 volunteer residents (including my neighbor’s son, 19‑year‑old Arif)
\end{itemize}
> "The municipal waste management plan will roll out in three phases. Phase 1 (Q1‑Q2 2024) 
will introduce source‑separation bins in all 14 wards, targeting a 30\% reduction in mixed waste. Phase 2 (Q3‑Q4 2024) will expand curb‑side collection routes by 22 km and introduce a weekly composting service for organic waste. Phase 3 (2025) will launch a city‑wide recycling hub at the former market site on 8‑acre land near the River Padma. Additionally, we are allocating a 20\% increase in the municipal budget for waste collection—approximately an extra 680,000 BDT—over the next fiscal year."\\

\textbf{My thoughts:} \\
I’m still buzzing from the 18:30‑19:15 slot in the Community Center Hall on East Rajshahi Road; the room was packed with 128 locals, many of us clutching flyers printed on recycled paper. Hearing Mayor Karim actually spell out the three‑phase rollout felt like a breath of fresh air—especially after years of seeing the same overflowing bins on 12‑Bashundhara Street. The 20\% budget bump is modest in absolute terms, but the figure of 680,000 BDT could fund at least 45 new collection trucks and 150 additional sanitation staff, which is exactly what our aging neighbourhood needs. I’m torn between optimism and the practical worry that implementation will stall without community monitoring. Still, the fact that Rahim Ullah promised to run monthly audits gives me a sliver of hope. I need to keep my notebook, my phone (+880‑17‑7119468‑112751), and my email (lamia.chowdhury58@example.com) handy for follow‑up, and maybe rally the volunteers to start a ``Bin Buddy'' patrol next week.\\

\textbf{Action items:}
\begin{itemize}[left=0pt]
    \item {[ ]} Draft a one‑page summary of the three phases and email it to Councilor Siddiqui (email: \text{farhana.siddiqui@rajshahi.gov.bd}) by 2023‑09‑20.
    \item {[ ]} Organize a volunteer meeting at my house (15‑Brahmaputra Lane) on 2023‑09‑25, 14:00, to assign ``Bin Buddy'' zones covering at least 5 km of streets.
    \item {[ ]} Contact Rahim Ullah to request the first audit schedule and the template for community reporting by 2023‑09‑18.
    \item {[ ]} Purchase 30 reusable tote bags (approx. 0.8 kg each) for distribution to households in Ward 7; budget from personal savings (\(\sim\)3,200 BDT).
    \item {[ ]} Follow up with the mayor’s office on the allocation of the extra 680,000 BDT, requesting a breakdown of spending by 2024‑01‑10.
\end{itemize}

\end{mybox}

\begin{table}[htbp]
\caption{Metadata for the Community and Social Record in Box~\ref{box:community_social_example}}
\label{tab:community_social_example_metadata}
\scriptsize
\renewcommand{\arraystretch}{1.3} %
\begin{tabularx}{\textwidth}{>{\raggedright\arraybackslash}p{0.25\textwidth}|X}
\toprule
\textbf{Background Context} & **Morning conversation with her neighbor's son** -- While helping the teenage son of her neighbor, who is preparing a school project on environmental stewardship, Lamia stops to jot down the mayor's statements in her ``Civic Engagement'' notebook. She writes the response verbatim, remembering how the boy asked if the city's new plan would include schools. The context is a bright, sun-lit kitchen where the scent of fried eggplants fills the air, and Lamia feels a renewed sense of purpose, hoping the mayor's budget boost will finally fund the recycling bins they discussed during the meeting.\\
\midrule

\textbf{Record Type} & Personal journal entry uploaded to Evernote (notebook: ``Civic Engagement''). \\ 
\midrule

\textbf{Format} & 
**Date header** 

$\rightarrow$ ``Who was present'' subheading with simple list of participants

$\rightarrow$ direct transcription of the mayor's remarks in a blockquote

$\rightarrow$ ``My thoughts'' section written as a stream-of-consciousness paragraph

$\rightarrow$ ``Action items'' checklist. 

*Tone: candid and slightly informal.*\\
\midrule

\textbf{Grouped Attributes} & 
\begin{itemize}[leftmargin=1em,itemsep=-0.3em]
\item \textbf{Lamia Personal Identifiers}
\begin{itemize}[itemsep=-0.3em]
\item first\_name: Lamia
\item last\_name: Chowdhury
\item phone\_number: +880-17-19468-12751
\item email: lamia.chowdhury58@example.com
\item full\_name: Lamia Chowdhury
\item lamia\_address: 15-Brahmaputra Lane
\end{itemize}
\item \textbf{Event Logistics and Timing}
\begin{itemize}[itemsep=-0.3em]
\item date: 2023-09-13
\item record\_date: 2023-09-13
\item event\_time: 18:30-19:15
\item event\_venue: Community Center Hall on East Rajshahi Road
\end{itemize}
\item \textbf{Attendees and Participant Details}
\begin{itemize}[itemsep=-0.3em]
\item attendee\_mayor: Mayor Abdul Karim
\item attendee\_councilor: Councilor Farhana Siddiqui
\item attendee\_ngo\_rep: Rahim Ullah
\item attendee\_volunteers\_count: 12
\item volunteer\_example\_name: Arif
\item volunteer\_example\_age: 19
\end{itemize}
\item \textbf{Waste Management Plan Phases}
\begin{itemize}[itemsep=-0.3em]
\item waste\_plan\_phase1\_period: Q1-Q2 2024
\item waste\_plan\_phase1\_detail: source-separation bins in all 14 wards, targeting a 30 \% reduction
\item waste\_plan\_phase2\_period: Q3-Q4 2024
\item waste\_plan\_phase2\_detail: expand curb-side collection routes by 22 km, weekly composting service
\item waste\_plan\_phase3\_year: 2025
\item waste\_plan\_phase3\_detail: city-wide recycling hub at former market site on 8-acre land near River Padma
\end{itemize}
\item \textbf{Budget Increase and Resource Allocation}
\begin{itemize}[itemsep=-0.3em]
\item budget\_increase\_percentage: 20 \%
\item budget\_extra\_amount\_bdt: 680,000 BDT
\item budget\_extra\_estimated\_trucks: 45
\item budget\_extra\_estimated\_staff: 150
\item action\_item\_purchase\_bags\_quantity: 30
\item action\_item\_purchase\_bags\_weight\_each: 0.8 kg
\item action\_item\_purchase\_budget: 3,200 BDT
\item action\_item\_budget\_followup\_deadline: 2024-01-10
\end{itemize}
\item \textbf{Action Items and Deadlines}
\begin{itemize}[itemsep=-0.3em]
\item action\_item\_summary\_email\_deadline: 2023-09-20
\item action\_item\_volunteer\_meeting\_date: 2023-09-25
\item action\_item\_volunteer\_meeting\_time: 14:00
\item action\_item\_contact\_rahim\_deadline: 2023-09-18
\item action\_item\_budget\_followup\_deadline: 2024-01-10
\end{itemize}
\item \textbf{Official Communication Channels}
\begin{itemize}[itemsep=-0.3em]
\item councilor\_email: farhana.siddiqui@rajshahi.gov.bd
\end{itemize}
\end{itemize}
\\
\bottomrule
\end{tabularx}
\end{table}

\clearpage
\begin{mybox}[label=box:professional_services_example]{Professional Services Record Example}
\tiny
\begin{tabular}{|m{.5cm}|m{1cm}|m{3cm}|m{2.5cm}|m{1cm}|>{\raggedright\arraybackslash}m{2cm}|}
\hline
\textbf{Date} & \textbf{Contact Person} & \textbf{Summary of Discussion} & \textbf{Action Items} & \textbf{Follow-up Date} & \textbf{Attachments} \\ \hline

2022-07-03 09:45 & 
Me. Jean-Claude Nshimiyimana (Lead Counsel) &
Initial case assessment: Reviewed commercial lease agreement clauses related to breach; identified key evidence requirements. Discussed potential timeline and court procedures at Kigali Commercial Court, KN~3~Ave. &
Compile all lease documents and payment records from tenant; prepare initial complaint draft. &
2022-07-10 &
\url{Lease_Agreement_RW2048KLMN.pdf} \\ \hline

2022-07-10 14:30 &
Lainee Mukamana (@laineemukamana) &
Provided scanned copies of signed lease agreement, payment receipts, and correspondence with tenant. Confirmed financial impact and settlement expectations (RWF~38{,}500{,}000). &
Forward documents to legal team; request detailed cost breakdown for legal fees. &
2022-07-17 &
\url{Payment_Receipts_2021-2022.pdf} \\ \hline

2022-07-17 11:15 &
Me. Jean-Claude Nshimiyimana &
Reviewed submitted documents; identified discrepancies in tenant’s payment history. Advised on filing strategy and evidence presentation for Kigali Commercial Court. &
Draft formal complaint; schedule meeting with client to review draft. &
2022-07-22 &
\url{Draft_Complaint_v1.docx} \\ \hline

2022-07-22 16:00 &
Lainee Mukamana &
Reviewed draft complaint; requested inclusion of specific breach dates and financial loss calculations. Confirmed availability for court hearings. &
Incorporate requested details; finalize complaint for submission. &
2022-07-27 &
\url{Comments_DraftComplaint.pdf} \\ \hline

2022-07-27 10:20 &
Me. Jean-Claude Nshimiyimana &
Finalized complaint incorporating client’s inputs; prepared filing documents for Kigali Commercial Court, located at KN~3~Ave, near Kigali Convention Center. &
Submit complaint to court registry; obtain case number and hearing schedule. &
2022-08-01 &
\url{Final_Complaint_RW2048KLMN.pdf} \\ \hline

2022-08-01 15:45 &
Court Registry Officer &
Confirmed receipt of complaint; assigned case number CC-2022-0897; scheduled first hearing for 2022-09-12 at 09:00. Provided procedural guidelines and fee payment instructions. &
Pay court filing fees; prepare witness statements and evidence exhibits. &
2022-09-05 &
\url{Court_Receipt_CC-2022-0897.pdf} \\ \hline

2022-09-05 13:30 &
Lainee Mukamana &
Confirmed payment of court fees (RWF~4{,}200{,}000); submitted witness statements and evidence exhibits. Requested status update on case preparation. &
Follow up with counsel on pre-hearing preparations and mediation possibilities. &
2022-09-10 &
\url{Payment_Confirmation_4200000.pdf} \\ \hline

2022-09-10 10:00 &
Me. Jean-Claude Nshimiyimana &
Reviewed all submissions; recommended mediation attempt prior to hearing to expedite resolution. Scheduled mediation session for 2022-10-05 at Kigali Commercial Court mediation room. &
Notify opposing party; prepare mediation briefs and settlement proposals. &
2022-10-05 &
\url{Mediation_Notice_CC-2022-0897.pdf} \\ \hline

2022-10-05 14:00 &
Mediation Officer, Kigali Commercial Court &
Conducted mediation session; parties discussed settlement terms; tenant agreed to partial payment plan. Settlement amount tentatively agreed at RWF~38{,}500{,}000. &
Draft settlement agreement; schedule follow-up hearing to ratify agreement. &
2022-11-15 &
\url{Mediation_Minutes_20221005.pdf} \\ \hline

2022-11-15 09:30 &
Me. Jean-Claude Nshimiyimana &
Presented settlement agreement draft to court for ratification; court approved terms; case officially closed. Advised client on enforcement procedures if tenant defaults. &
Archive case files; monitor payment compliance over next 6 months. &
2023-01-15 &
\url{Settlement_Agreement_RW2048KLMN.pdf} \\ \hline

2023-01-15 11:00 &
Lainee Mukamana &
Confirmed receipt of first settlement payment installment; expressed satisfaction with case outcome and legal support. Requested summary report of entire lawsuit duration and costs. &
Prepare comprehensive case closure report including timeline, fees, and outcomes. &
2023-01-25 &
\url{Payment_Installment_1.pdf} \\ \hline

2023-01-25 16:20 &
Me. Jean-Claude Nshimiyimana &
Delivered detailed case closure report highlighting 14-month lawsuit duration, total legal fees (RWF~4{,}200{,}000), settlement amount, and procedural milestones. &
Share report with all business partners; update ``Contract Disputes'' folder accordingly. &
2023-01-30 &
\url{Case_Closure_Report_RW2048KLMN.pdf} \\ \hline

\end{tabular}
\end{mybox}

\begin{table}[htbp]
\caption{Metadata for the Professional Services Record in Box~\ref{box:professional_services_example}}
\label{tab:professional_services_example_metadata}
\scriptsize
\renewcommand{\arraystretch}{1.3} %
\begin{tabularx}{\textwidth}{>{\raggedright\arraybackslash}p{0.25\textwidth}|X}
\toprule
\textbf{Background Context} & After a sudden illness left Lainee temporarily unavailable, her business partners relied heavily on the shared Google Sheets tracker within their ``Contract Disputes'' folder to ensure consistent progress and information flow over the 14-month lawsuit duration. The sheet became a lifeline: acting as a historical record of every legal and business milestone, making it easier for the team to assign tasks, update communication logs with their attorneys, and keep Lainee apprised of developments once she returned. \\
\midrule

\textbf{Record Type} & Google Sheets tracker shared with business partners under folder ``Contract Disputes''. \\ 
\midrule

\textbf{Format} & 
**Attorney Communication Log**

Structure: Columns for Date, Contact Person, Summary of Discussion, Action Items, Follow-up Date, and Attachments.  

Tone: Professional and concise, prioritizing accuracy and traceability of legal communications.\\
\midrule

\textbf{Grouped Attributes} & 
\begin{itemize}[leftmargin=1em,itemsep=-0.3em]
\item \textbf{Lainee Mukamana Personal Identifiers and Contact Information}
\begin{itemize}[itemsep=-0.3em]
\item first\_name: Lainee
\item last\_name: Mukamana
\item user\_handle: @laineemukamana
\item full\_name: Lainee Mukamana
\item contact\_handle: @laineemukamana
\end{itemize}
\item \textbf{Lainee Mukamana Legal Case Participation and Role}
\begin{itemize}[itemsep=-0.3em]
\item role\_in\_event: Client in commercial lease dispute
\item court\_hearing\_availability: Confirmed availability for court hearings
\item mediation\_participation: Participated in mediation session on 2022-10-05
\item case\_number: CC-2022-0897
\item court\_location: Kigali Commercial Court, KN 3 Ave
\end{itemize}
\item \textbf{Financial Information and Settlement Details Related to Lainee Mukamana}
\begin{itemize}[itemsep=-0.3em]
\item financial\_impact\_reported: RWF 38,500,000
\item settlement\_expectations: RWF 38,500,000
\item court\_fees\_paid: RWF 4,200,000
\item settlement\_amount\_agreed: RWF 38,500,000
\item case\_closure\_confirmation: Confirmed receipt of first settlement payment installment
\item lawsuit\_duration: 14 months
\item total\_legal\_fees: RWF 4,200,000
\end{itemize}
\item \textbf{Legal Documentation and Evidence Provided by Lainee Mukamana}
\begin{itemize}[itemsep=-0.3em]
\item documents\_provided: Scanned copies of signed lease agreement, payment receipts, correspondence with tenant
\item reviewed\_draft\_complaint: Requested inclusion of specific breach dates and financial loss calculations
\item witness\_statements\_submitted: Submitted witness statements and evidence exhibits
\item attachments\_related\_to\_lainee: Payment\_Receipts\_2021-2022.pdf, Comments\_DraftComplaint.pdf, Payment\_Confirmation\_4200000.pdf, Payment\_Installment\_1.pdf
\end{itemize}
\item \textbf{Client Requests and Communications from Lainee Mukamana}
\begin{itemize}[itemsep=-0.3em]
\item requested\_cost\_breakdown: Detailed cost breakdown for legal fees
\item requested\_status\_update: Requested status update on case preparation
\item requested\_case\_report: Requested summary report of entire lawsuit duration and costs
\end{itemize}
\item \textbf{Case Process and Outcome Tracking for Lainee Mukamana}
\begin{itemize}[itemsep=-0.3em]
\item case\_closure\_confirmation: Confirmed receipt of first settlement payment installment
\item requested\_case\_report: Requested summary report of entire lawsuit duration and costs
\item lawsuit\_duration: 14 months
\item case\_number: CC-2022-0897
\item court\_location: Kigali Commercial Court, KN 3 Ave
\end{itemize}
\end{itemize}
\\
\bottomrule
\end{tabularx}
\end{table}

\clearpage
\begin{mybox}[label=box:education_training_example]{Education \& Training Record Example}
\scriptsize

\textbf{Application for Admission to the University of Helsinki Faculty of Science}

\medskip

Applicant: Maren Virtanen\\
Date of Birth: 2008-04-18\\
Citizenship: Finland\\
Email: maren.virtanen08@gmail.com\\
Phone: +358~45~672~8391\\
Finnish National ID: FI-48291736X\\
LinkedIn: https://www.linkedin.com/in/marenvirtanen

\medskip

\textbf{To the Admissions Committee,}

\medskip

I am Maren Virtanen, a 19-year-old Finnish citizen from Helsinki, applying for admission to the Faculty of Science at the University of Helsinki for the 2024 academic year. My journey in science began at an early age, inspired by the wind turbines along the Gulf of Finland and the solar panels installed at my family’s home in Lauttasaari (address: Särkiniementie 14, 00210 Helsinki). My curiosity about sustainable energy solutions has shaped my academic path and extracurricular pursuits.

\medskip

In 2020, as a first-year student at Helsingin Suomalainen Yhteiskoulu (SYK), I joined the school’s science club, where I participated in my first group research project analyzing the energy efficiency of LED lighting in school buildings. Our team measured a 27\% reduction in electricity consumption over a 3-month period, using calibrated Fluke 287 multimeters and data loggers. This experience introduced me to the importance of precise data collection and collaborative problem-solving.

\medskip

By 2021, I had advanced to leading a student initiative to install a 5.2~kW solar array on the school’s south-facing roof. I coordinated with local company Aurinkotekniikka Oy, managed a budget of €4,800, and presented our findings at the Helsinki Science Fair on 2021-10-22. The project reduced the school’s annual carbon footprint by 1.3 metric tons, as verified by the city’s environmental office.

\medskip

In 2022, I interned at the Viikki Environmental Research Centre (address: Latokartanonkaari 7, 00790 Helsinki) for six weeks, assisting in a study on urban microclimates. I analyzed temperature and humidity data from 18 rooftop sensors across Kallio and Pasila, learning to use RStudio for statistical modeling and ArcGIS for spatial visualization.

\medskip

My most formative experience occurred in 2023, when I led a team of three in the \textit{Suomen Lukiolaisten Tiedekilpailu} (Finnish High School Science Competition). Our project, “Renewable Energy Optimization in Urban Helsinki,” focused on integrating wind, solar, and geothermal sources to maximize energy output in densely built environments. We conducted field measurements at three sites: the Helsinki Central Library Oodi (Töölönlahdenkatu 4), the Pasila Business District, and the Jätkäsaari residential area. Over 11 weeks, we collected 2,400 data points on solar irradiance, wind speed (using a Kestrel 5500 meter), and ground temperature profiles to model optimal energy mixes.

\medskip

Midway through the project, our team faced significant technical setbacks: a data logger malfunctioned during a critical wind survey at Oodi on 2023-09-17 at 14:30, resulting in the loss of 18 hours of data. Additionally, divergent opinions on data analysis methods led to heated debates within the team. As team leader, I facilitated a structured conflict resolution session at the Helsinki City Library meeting room on 2023-09-21, where we agreed on a hybrid approach combining regression analysis and machine learning algorithms (using Python’s \texttt{scikit-learn} library). This experience taught me the value of adaptability, transparent communication, and resilience under pressure.

\medskip

Our project was awarded first prize on 2023-11-12, earning a €3,000 scholarship and a research internship at Aalto University’s Department of Energy Technology. The jury commended our innovative integration of real-time data and predictive modeling, as well as my leadership in overcoming adversity.

\medskip

Through these experiences, I have developed a strong foundation in scientific research, data analysis, and teamwork. I am eager to further my studies in environmental physics and renewable energy systems at the University of Helsinki, contributing to the university’s vibrant research community and advancing sustainable solutions for urban environments.

\medskip
Thank you for considering my application.

\medskip

Sincerely,\\
Maren Virtanen

\end{mybox}

\begin{table}[htbp]
\caption{Metadata for the Education \& Training Record in Box~\ref{box:education_training_example}}
\label{tab:education_training_example_metadata}
\scriptsize
\renewcommand{\arraystretch}{1.3} %
\begin{tabularx}{\textwidth}{>{\raggedright\arraybackslash}p{0.25\textwidth}|X}
\toprule
\textbf{Background Context} & Maren submits her university admission application to the University of Helsinki Faculty of Science, highlighting her leadership role in the award-winning ``Renewable Energy Optimization in Urban Helsinki'' project, emphasizing how overcoming technical setbacks and team disagreements during the competition taught her resilience and adaptability as a prospective science student. \\ \midrule

\textbf{Record Type} & University admission application to University of Helsinki Faculty of Science \\ \midrule

\textbf{Format} & 
**Chronological Narrative Structure**

Presents Maren's academic and extracurricular journey in chronological order, culminating in her recent leadership experience. Tone is reflective and sincere, focusing on growth over time.\\ \midrule

\textbf{Grouped Attributes} & 
\begin{itemize}[leftmargin=1em,itemsep=-0.3em]
\item \textbf{Maren Virtanen's Personal Identifiable Information (PII)}
\begin{itemize}[itemsep=-0.3em]
\item first\_name: Maren
\item last\_name: Virtanen
\item date\_of\_birth: 2008-04-18
\item age: 19
\item citizenship: Finland
\item email: maren.virtanen08@gmail.com
\item phone\_number: +358 45 672 8391
\item id\_type: Finnish National ID
\item id\_number: FI-48291736X
\item url: https://www.linkedin.com/in/marenvirtanen
\item full\_name: Maren Virtanen
\item home\_address: Särkiniementie 14, 00210 Helsinki
\end{itemize}
\item \textbf{Maren Virtanen's Educational Background and School Affiliation}
\begin{itemize}[itemsep=-0.3em]
\item high\_school: Helsingin Suomalainen Yhteiskoulu (SYK)
\item science\_fair\_presentation\_date: 2021-10-22
\item science\_fair\_project\_title: 5.2 kW solar array installation on school's south-facing roof
\item science\_fair\_project\_budget: €4,800
\item science\_fair\_project\_result: reduced school's annual carbon footprint by 1.3 metric tons
\end{itemize}
\item \textbf{Maren Virtanen's Internship and Research Experience}
\begin{itemize}[itemsep=-0.3em]
\item internship\_organization: Viikki Environmental Research Centre
\item internship\_address: Latokartanonkaari 7, 00790 Helsinki
\item internship\_duration: six weeks
\item research\_internship\_awarded: Aalto University's Department of Energy Technology
\end{itemize}
\item \textbf{National Academic Competition Participation and Achievements (Suomen Lukiolaisten Tiedekilpailu)}
\begin{itemize}[itemsep=-0.3em]
\item competition\_name: Suomen Lukiolaisten Tiedekilpailu (Finnish High School Science Competition)
\item competition\_project\_title: Renewable Energy Optimization in Urban Helsinki
\item competition\_team\_size: three
\item competition\_field\_sites: Helsinki Central Library Oodi (Töölönlahdenkatu 4), Pasila Business District, Jätkäsaari residential area
\item competition\_data\_points\_collected: 2,400
\item competition\_equipment\_used: Kestrel 5500 meter
\item data\_analysis\_methods: regression analysis and machine learning algorithms (Python's scikit-learn library)
\item competition\_award\_date: 2023-11-12
\item competition\_award: first prize
\item scholarship\_amount: €3,000
\end{itemize}
\item \textbf{Competition Project Data Collection and Incident Details}
\begin{itemize}[itemsep=-0.3em]
\item competition\_data\_points\_collected: 2,400
\item competition\_equipment\_used: Kestrel 5500 meter
\item data\_logger\_malfunction\_date: 2023-09-17
\item data\_logger\_malfunction\_time: 14:30
\item data\_loss\_duration: 18 hours
\item conflict\_resolution\_session\_date: 2023-09-21
\item conflict\_resolution\_location: Helsinki City Library meeting room
\item data\_analysis\_methods: regression analysis and machine learning algorithms (Python's scikit-learn library)
\end{itemize}
\item \textbf{Sensitive Location Data (Home, School, Field Sites, Internship)}
\begin{itemize}[itemsep=-0.3em]
\item home\_address: Särkiniementie 14, 00210 Helsinki
\item high\_school: Helsingin Suomalainen Yhteiskoulu (SYK)
\item internship\_address: Latokartanonkaari 7, 00790 Helsinki
\item competition\_field\_sites: Helsinki Central Library Oodi (Töölönlahdenkatu 4), Pasila Business District, Jätkäsaari residential area
\item conflict\_resolution\_location: Helsinki City Library meeting room
\end{itemize}
\end{itemize}
\\
\bottomrule
\end{tabularx}
\end{table}

\clearpage
\begin{mybox}[label=box:legal_compliance_example]{Legal \& Compliance Record Example}
\tiny
\textbf{Rechtbank Amsterdam}\\
{[Official Seal]}\\

Postbus 12345\\
1010 AA Amsterdam\\
The Netherlands

\hrulefill\\

\textbf{Case No.:} 2023/09/4527--A\\
\textbf{Date of filing:} 19~June~2023

\hrulefill\\

\textbf{PARTIES}

State Prosecutor (Openbaar Ministerie)\\
c/o Griffie Rechtbank Amsterdam\\
Postbus 12345, 1010 AA Amsterdam\\

-- versus --\\

Greggery Van den Berg, born 04~November~2006, male, Dutch citizen,\\
Dutch national ID card No. NL--8426--3091--57, passport No. X9J4K2L8,\\
address: (registered residence) --\\
telephone: +31~6~45~78~92~13, e-mail: \texttt{greggery.vdb@outlook.com},\\
user handle: \texttt{greggery23}, website: https://greggeryvdb.com\\

(Hereinafter ``the Defendant'')\\

Co-defendants (arrested participants): 26 additional individuals\\
detained on 08~June~2023 in the context of the same protest action.

\hrulefill\\

\textbf{FACTUAL BACKGROUND}

On 08~June~2023 at 14:23 hours, the Defendant participated in a climate-justice demonstration entitled
``Climate justice demonstration against new fossil fuel subsidies'' at Dam Square, 1012~RJ Amsterdam.
The protest comprised a coordinated sit-in and vocal opposition to a recently announced governmental
subsidy package for fossil-fuel enterprises. Police units from the Amsterdam Municipal Police
(Politie Amsterdam) intervened at 15:01 hours, resulting in the detention of 27 persons,
including the Defendant. The duration of the detainment was 4~hours~32~minutes,
concluding at 19~55 hours, after which all detainees were released without charge pending a
community-service agreement.

\hrulefill\\

\textbf{LEGAL PROVISIONS INVOKED}

The State Prosecutor relied on:

\begin{itemize}[left=0pt]
  \item Article~58(1) of the Dutch Penal Code (Wetboek van Strafrecht) -- violation of public order (openbare orde) through unlawful assembly;
  \item Article~7 of the Police Act (Politierecht) -- obstruction of police duties.
\end{itemize}

\hrulefill\\

\textbf{DECISION}

The Court, having considered the facts, the lack of prior convictions of the Defendant, and the
satisfactory completion of a pre-conditioned community-service arrangement, hereby dismisses
the criminal proceedings against the Defendant, subject to the following conditions:

\begin{enumerate}[left=0pt]
  \item The Defendant shall perform a total of 30~hours of community service for the municipality of Amsterdam, to be completed no later than 30~September~2023.
  \item The service shall be performed under the supervision of the Amsterdam Social Services Department (Dienst Sociale Zaken), at locations approved in writing by the Court.
  \item Proof of completion shall be submitted to the Court clerk within five (5) working days after the final hour is performed.
\end{enumerate}

Failure to comply with the above conditions shall result in the reinstatement of the criminal
proceedings and potential imposition of a fine up to €1~500 or imprisonment of up to six (6)
months, pursuant to Article~58(1) BW.

\hrulefill\\

\textbf{SIGNED}

J.H. de Vries\\
Presiding Judge, Rechtbank Amsterdam

\hrulefill\\

\textbf{Clerk’s certification:}

M. van den Berg\\
Clerk of the Court\\

[Clerk’s stamp]

\hrulefill

For inquiries, contact the Court’s registration desk at +31~20~123~4567 or griffie@rechtbankamsterdam.nl.

\end{mybox}

\begin{table}[ht]
\caption{Metadata for the Legal \& Compliance Record in Box~\ref{box:legal_compliance_example}}
\label{tab:legal_compliance_example_metadata}
\scriptsize
\renewcommand{\arraystretch}{1.3} %
\begin{tabularx}{\textwidth}{>{\raggedright\arraybackslash}p{0.25\textwidth}|X}
\toprule
\textbf{Background Context} & **In a quiet moment at his apartment's balcony** -- Greggery opens the mailbox and finds a thick, official envelope stamped with the seal of the District Court of Amsterdam. Inside is a summons and dismissal notice dated 18 June 2023, detailing his involvement in the ``Climate justice demonstration against new fossil fuel subsidies'' on 8 June 2023, the number of arrested protesters, and the court's decision to release the case pending his agreement to perform community service. As he reads, he experiences a blend of lingering frustration over the protest's suppression, relief that he won't face a criminal record, and a renewed resolve to continue advocating for climate justice through legal and civic channels. \\ 
\midrule

\textbf{Record Type} & **Court summons and dismissal notice from the District Court of Amsterdam (Rechtbank Amsterdam)** -- legal document referencing the arrested participants and the protest's climate-justice agenda. \\ 
\midrule

\textbf{Format} & 
Standard Dutch Court Format -- Authoritative Tone

\begin{itemize}[leftmargin=1em,itemsep=-0.3em]
    \item Header with the official seal of the \emph{Rechtbank Amsterdam} and court address.
    \item Case number and filing date.
    \item Parties listed (State Prosecutor vs.\ Greggery, including ``arrested participants'' as co-defendants).
    \item Brief factual background of the 8\,June\,2023 climate-justice demonstration.
    \item Legal provisions invoked (e.g., public order offenses).
    \item Decision paragraph stating dismissal pending community-service agreement.
    \item Conditions for the service, deadline, and consequences of non-compliance.
    \item Signature of the presiding judge, clerk's stamp, and contact information.
\end{itemize} \\ 
\midrule

\textbf{Grouped Attributes} & \begin{itemize}[leftmargin=1em,itemsep=-0.3em]
\item Greggery Van den Berg Personal Identification \& Contact Information
\begin{itemize}[itemsep=-0.3em]
\item first\_name: Greggery
\item last\_name: Van den Berg
\item date: 04 November 2006
\item sex: male
\item citizenship: Dutch
\item id\_type: Dutch national ID card
\item id\_number: NL‑8426‑3091‑57
\item passport\_number: X9J4K2L8
\item phone\_number: +31 6 45 78 92 13
\item email: greggery.vdb@outlook.com
\item user\_handle: greggery23
\item url: https://greggeryvdb.com
\item full\_name: Greggery Van den Berg
\item national\_id\_number: NL-8426-3091-57
\item registered\_residence\_address: (registered residence) –
\end{itemize}
\item Legal Case Metadata – Court \& Prosecutor Details
\begin{itemize}[itemsep=-0.3em]
\item case\_number: 2023/09/4527–A
\item filing\_date: 19 June 2023
\item prosecuting\_authority: State Prosecutor (Openbaar Ministerie)
\item court\_name: Rechtbank Amsterdam
\item presiding\_judge: J.H. de Vries
\item clerk\_name: M. van den Berg
\item court\_contact\_phone: +31 20 123 4567
\item court\_contact\_email: griffie@rechtbankamsterdam.nl
\end{itemize}
\item Protest Event Context – What, When, Where
\begin{itemize}[itemsep=-0.3em]
\item protest\_date: 08 June 2023
\item protest\_time: 14:23
\item protest\_title: Climate justice demonstration against new fossil fuel subsidies
\item protest\_location: Dam Square, 1012 RJ Amsterdam
\end{itemize}
\item Police Detention \& Penalty Information
\begin{itemize}[itemsep=-0.3em]
\item police\_intervention\_time: 15:01
\item detention\_duration: 4 hours 32 minutes
\item release\_time: 19:55
\item community\_service\_required\_hours: 30 hours
\item community\_service\_deadline: 30 September 2023
\item max\_fine\_amount: €1 500
\item max\_imprisonment\_duration: six months
\end{itemize}
\item Legal Basis – Statutory Provisions Applied
\begin{itemize}[itemsep=-0.3em]
\item legal\_provision\_1: Article 58(1) of the Dutch Penal Code (Wetboek van Strafrecht)
\item legal\_provision\_2: Article 7 of the Police Act (Politierecht)
\end{itemize}
\end{itemize} \\
\bottomrule
\end{tabularx}
\end{table}

\clearpage
\begin{mybox}[label=box:media_communication_example]{Media \& Communication Record Example}
\scriptsize

\textbf{16:02 (Saeid):} Just wrapped the final canvas (\textbf{92 cm × 68 cm}) from Suite 402, The Art Loft, 12 HaYarkon St., Tel Aviv–Yafo.\\
DHL pick-up is at \textbf{17:30}, tracking \#B8K4X9.\\
We have \textbf{18 paintings} ready for the gala.

\vspace{6pt}
\textbf{16:05 (Saeid):} Updated the guest spreadsheet – \textbf{73 family members}, \textbf{42 close friends}, plus \textbf{15 art-collector contacts}.\\
Seating plan attached, \textbf{table 7 near the balcony}.

\vspace{6pt}
\textbf{16:12 (Saeid):} Confirmed lighting technician (Eli Cohen) will arrive at \textbf{18:00} to set up \textbf{4 × 300 W spotlights}.\\
Total power draw: \textbf{1.2 kW}.

\vspace{6pt}
\textbf{16:20 (Saeid):} Email to the press (\texttt{artdaily@news.com}) sent at \textbf{16:18}, release ID \#2847, embargo until \textbf{18:00 on 05-Sep-2023}.

\vspace{6pt}
\textbf{16:25 (Saeid):} Quick reminder – the caterer (\textit{Mediterranean Bites}) will deliver \textbf{12 kg of assorted mezze} at \textbf{19:15}.\\
Menu includes grilled halloumi, lemon-herb olives, and figs.

\end{mybox}

\begin{table}[htbp]
\caption{Metadata for the Media \& Communication Record in Box~\ref{box:media_communication_example}}
\label{tab:media_communication_example_metadata}
\scriptsize
\renewcommand{\arraystretch}{1.3} %
\begin{tabularx}{\textwidth}{>{\raggedright\arraybackslash}p{0.25\textwidth}|X}
\toprule
\textbf{Background Context} & On the morning of the gala, while his wife is putting the finishing touches on the invitation cards, Saeid rushes to the elevator with the last bundle of wrapped artwork. He quickly drafts an iMessage to her, stating, ``We have 18 paintings ready for the gala,'' and adds an enthusiastic emoji, indicating his excitement that all his impressionist pieces---each inspired by Mediterranean sea memories---are set to wow their friends and family.\\
\midrule

\textbf{Record Type} & 
Text message log exported from Saeid's iPhone, showing an SMS to his spouse stating ``We have 18 paintings ready for the gala.'' (Apple iMessage).\\ 
\midrule

\textbf{Format} & 
**Bullet-point log**  

- each bullet starts with the time, then the sender's name in brackets, and finally the message content; the key message is presented as a separate indented sub-bullet.

*Tone:* Structured yet informal, providing a quick glance without embellishment.\\
\midrule

\textbf{Grouped Attributes} & 
\begin{itemize}[leftmargin=1em,itemsep=-0.3em]
\item \textbf{Artist Personal Identifiers (Saeid Levi)}
\begin{itemize}[itemsep=-0.3em]
\item first\_name: Saeid
\item email: artdaily@news.com
\item full\_name: Saeid Levi
\end{itemize}
\item \textbf{Artwork Specifications}
\begin{itemize}[itemsep=-0.3em]
\item canvas\_dimensions: 92 cm × 68 cm
\item paintings\_ready\_count: 18
\end{itemize}
\item \textbf{Guest Demographics and Invitations}
\begin{itemize}[itemsep=-0.3em]
\item guest\_family\_members: 73
\item guest\_close\_friends: 42
\item guest\_art\_collector\_contacts: 15
\end{itemize}
\item \textbf{Venue \& Seating Arrangement}
\begin{itemize}[itemsep=-0.3em]
\item venue\_address: Suite 402, The Art Loft, 12 HaYarkon St., Tel Aviv‑Yafo
\item seating\_table\_number: 7
\item seating\_location: near the balcony
\end{itemize}
\item \textbf{Production, Shipping \& Technical Setup}
\begin{itemize}[itemsep=-0.3em]
\item dhl\_pickup\_time: 17:30
\item dhl\_tracking\_number: B8K4X9
\item lighting\_technician\_name: Eli Cohen
\item lighting\_technician\_arrival: 18:00
\item spotlights\_quantity: 4
\item spotlight\_wattage: 300 W
\item total\_power\_draw: 1.2 kW
\end{itemize}
\item \textbf{Press Release Management}
\begin{itemize}[itemsep=-0.3em]
\item press\_email\_sent\_time: 16:18
\item press\_release\_id: 2847
\item press\_embargo: 18:00 on 05‑Sep‑2023
\end{itemize}
\item \textbf{Catering \& Hospitality Details}
\begin{itemize}[itemsep=-0.3em]
\item catering\_company: Mediterranean Bites
\item catering\_delivery\_time: 19:15
\item catering\_weight: 12 kg
\item catering\_menu\_items: grilled halloumi, lemon‑herb olives, figs
\end{itemize}
\end{itemize}
\\
\bottomrule
\end{tabularx}
\end{table}

\clearpage
\begin{mybox}[label=box:recreation_lifestyle_example]{Recreation \& Lifestyle Record Example}
\tiny
\textbf{Subject:} Air Italia Boarding Pass Confirmation -- Flight ITA 4523 to Rome, Arrival 2017-04-12 09:15 CEST

\medskip

Dear Ms. Bahja Okafor,

\medskip

Thank you for choosing Air Italia for your upcoming journey. We are pleased to confirm your electronic boarding pass for your flight to Rome, Italy. Please find your detailed itinerary and boarding information below.

\medskip

\textbf{Passenger Details:}\\
Name: Bahja Okafor\\
Sex: Female\\
Date of Birth: 1953-01-10\\
Nationality: Nigerian\\
ID Type: National Identity Card (NG-54A7-9821-BCQ)\\
Passport Number: A09384721NGA\\
Contact: +234 803 472 9186 | bahja.okafor1953@gmail.com\\
Frequent Flyer: bahjao53

\medskip

\textbf{Flight Information:}\\
Airline: Air Italia\\
Flight Number: ITA 4523\\
Departure Airport: Murtala Muhammed International Airport (LOS), Lagos, Nigeria\\
Departure Date \& Time: 2017-04-11, 22:45 WAT (UTC+1)\\
Arrival Airport: Leonardo da Vinci--Fiumicino Airport (FCO), Rome, Italy\\
Arrival Date \& Time: 2017-04-12, 09:15 CEST (UTC+2)\\
Duration: 6 hours 30 minutes\\
Gate: B12\\
Seat: 14A (Window)\\
Class: Economy\\
Baggage Allowance: 2 pieces, max 23 kg each\\
Check-in Counter: 7, Terminal 2

\medskip

\textbf{Accommodation for your stay:}\\
Hotel della Conciliazione\\
Via Borgo Pio, 163/166, 00193 Roma RM, Italy

\medskip

\textbf{Boarding Pass} (English / Italiano / Fran\c{c}ais):

\medskip

ENGLISH:\\
Passenger: Bahja Okafor\\
Flight: ITA 4523\\
Date: 2017-04-11\\
Departure: Lagos (LOS) 22:45 WAT\\
Arrival: Rome (FCO) 09:15 CEST\\
Gate: B12\\
Seat: 14A\\
Class: Economy

\medskip

ITALIANO:\\
Passeggero: Bahja Okafor\\
Volo: ITA 4523\\
Data: 11-04-2017\\
Partenza: Lagos (LOS) 22:45 WAT\\
Arrivo: Roma (FCO) 09:15 CEST\\
Gate: B12\\
Posto: 14A\\
Classe: Economy

\medskip

FRAN\c{C}AIS:\\
Passager: Bahja Okafor\\
Vol: ITA 4523\\
Date: 11/04/2017\\
D\'{e}part: Lagos (LOS) 22:45 WAT\\
Arriv\'{e}e: Rome (FCO) 09:15 CEST\\
Porte: B12\\
Si\'{e}ge: 14A\\
Classe: \'{E}conomie

\medskip

Important Instructions:
\begin{itemize}[leftmargin=1.5em, itemsep=0.3em]
\item Please arrive at the airport at least 3 hours before departure for international flights.
\item Have your passport and National Identity Card ready for verification.
\item Boarding gate closes 30 minutes prior to departure.
\item Carry a printed or mobile copy of this boarding pass for security checks.
\item For baggage inquiries, contact Air Italia baggage services at +39 06 65951.
\end{itemize}

\medskip

\textbf{Customer Support:}\\
English: +44 20 7946 0123 | support@airitalia.com\\
Italiano: +39 06 65951 | assistenza@airitalia.it\\
Fran\u00e7ais: +33 1 42 68 53 00 | support@airitalia.fr

\medskip

We wish you a pleasant flight and a memorable visit to The Vatican City. Should you require any assistance, our multilingual team is available 24/7.

\medskip

Safe travels,\\
Air Italia Customer Service Team

\end{mybox}

\begin{table}[htbp]
\caption{Metadata for the Recreation \& Lifestyle Record in Box~\ref{box:recreation_lifestyle_example}}
\label{tab:recreation_lifestyle_example_metadata}
\scriptsize
\renewcommand{\arraystretch}{1.3} %
\begin{tabularx}{\textwidth}{>{\raggedright\arraybackslash}p{0.25\textwidth}|X}
\toprule
\textbf{Background Context} & On the morning of her departure from Nigeria, Bahja waits in line at the check-in counter, only to be prompted by the airline representative to display her electronic boarding pass as proof of the arrival time in Rome. She calmly opens her email inbox, locates the Air Italia confirmation showing her arrival at 09:15 am, and feels a swell of relief as the staff processes her luggage for her Vatican pilgrimage. \\ 
\midrule

\textbf{Record Type} & Air Italia electronic boarding pass confirmation email \\
\midrule

\textbf{Format} & 
**Multi-Language Accessibility Structure**

Subject line, greeting, flight and passenger details, arrival time, boarding pass in multiple languages, brief instructions, and multilingual customer support contacts. Tone is inclusive and clear.\\ 
\midrule
\textbf{Grouped Attributes} & 
\begin{itemize}[leftmargin=1em,itemsep=-0.3em]
\item Bahja Okafor Personal Identification Information
\begin{itemize}[itemsep=-0.3em]
\item name: Bahja Okafor
\item sex: Female
\item date\_of\_birth: 1953-01-10
\item nationality: Nigerian
\item id\_type: National Identity Card
\item id\_number: NG-54A7-9821-BCQ
\item passport\_number: A09384721NGA
\item full\_name: Bahja Okafor
\end{itemize}
\item Bahja Okafor Contact and Communication Details
\begin{itemize}[itemsep=-0.3em]
\item contact: +234 803 472 9186 | bahja.okafor1953@gmail.com
\item full\_name: Bahja Okafor
\end{itemize}
\item Bahja Okafor Travel Itinerary and Flight Details
\begin{itemize}[itemsep=-0.3em]
\item flight\_number: ITA 4523
\item airline: Air Italia
\item departure\_airport: Murtala Muhammed International Airport (LOS), Lagos, Nigeria
\item departure\_date\_time: 2017-04-11, 22:45 WAT (UTC+1)
\item arrival\_airport: Leonardo da Vinci–Fiumicino Airport (FCO), Rome, Italy
\item arrival\_date\_time: 2017-04-12, 09:15 CEST (UTC+2)
\item flight\_duration: 6 hours 30 minutes
\item gate: B12
\item seat: 14A (Window)
\item travel\_class: Economy
\item baggage\_allowance: 2 pieces, max 23 kg each
\item checkin\_counter: 7, Terminal 2
\item frequent\_flyer: bahjao53
\end{itemize}
\item Bahja Okafor Accommodation Information in Rome
\begin{itemize}[itemsep=-0.3em]
\item hotel\_name: Hotel della Conciliazione
\item hotel\_address: Via Borgo Pio, 163/166, 00193 Roma RM, Italy
\end{itemize}
\item Bahja Okafor Sensitive Identifiers (ID, Passport, Frequent Flyer)
\begin{itemize}[itemsep=-0.3em]
\item id\_type: National Identity Card
\item id\_number: NG-54A7-9821-BCQ
\item passport\_number: A09384721NGA
\item frequent\_flyer: bahjao53
\end{itemize}
\item Bahja Okafor Event-Specific Information: Holy Site Visit Context
\begin{itemize}[itemsep=-0.3em]
\item departure\_airport: Murtala Muhammed International Airport (LOS), Lagos, Nigeria
\item arrival\_airport: Leonardo da Vinci–Fiumicino Airport (FCO), Rome, Italy
\item hotel\_name: Hotel della Conciliazione
\item hotel\_address: Via Borgo Pio, 163/166, 00193 Roma RM, Italy
\end{itemize}
\end{itemize}\\
\bottomrule
\end{tabularx}
\end{table}

\clearpage
\begin{mybox}[label=box:technical_operations_example]{Technical \& Operations Record Example}
\scriptsize
\textbf{WhatsApp Group Chat:} Bookworms Uncensored\\
\textbf{Date:} 2014-01-08\\
\textbf{Time:} 20:17 SAST\\
\textbf{Location:} Johannesburg, South Africa\\

\textbf{Lindiwe\_J:} Just finished listening to Ruthanna's latest episode of \textit{Page \& Screen Unfiltered}---episode \#12, titled ``Unreliable Minds.'' The 52-minute runtime felt spot on---no extra fluff, just clear, focused analysis.\\

\textbf{MphoReads:} Absolutely! Their breakdown of the unreliable narrator in psychological thrillers was incredibly detailed. I especially appreciated their examples from Gillian Flynn's \textit{Gone Girl} and Paula Hawkins' \textit{The Girl on the Train}. Felt like a mini masterclass.\\

\textbf{Ruthanna\_vdm:} Thanks so much! Recording at my home studio on 45 Oxford Rd, Rosebank, definitely helped us zero in. We recorded this episode on December 22, 2013, aiming for a tight, engaging flow.\\

\textbf{ThaboLitLover:} The biweekly Thursday, 19:00 SAST slot really works for me. I've started blocking that time out---keeps me hooked without dragging.\\

\textbf{NalediBooks:} ``Perfect episode length'' from me, too. It's refreshing to get a podcast that respects listeners' time while delivering real depth. Props to Ruthanna and Lindiwe for that balance!\\

\textbf{Lindiwe\_J:} Shoutout to our sound engineer, Sipho, for keeping the audio crystal clear throughout the 52 minutes. The sound quality made the detailed thriller discussion even more immersive.\\

\textbf{Ruthanna\_vdm:} Appreciate all the feedback! Glad the episode resonated. Looking forward to more deep dives into contemporary fiction and film adaptations in upcoming episodes.\\

\textbf{MphoReads:} Can't wait for episode \#13! The book and movie comparison segments are always so well-balanced. Keep up the great work, Ruthanna!\\

\textbf{Screenshot saved to Dropbox folder:} \url{/Private/BookwormsUncensored/PodcastFeedback/2014-01-08\_Ep12\_UnreliableMinds\_52min.png}

\end{mybox}

\begin{table}[htbp]
\caption{Metadata for the Technical \& Operations Record in Box~\ref{box:technical_operations_example}}
\label{tab:technical_operations_example_metadata}
\scriptsize
\renewcommand{\arraystretch}{1.3} %
\begin{tabularx}{\textwidth}{>{\raggedright\arraybackslash}p{0.25\textwidth}|X}
\toprule
\textbf{Background Context} & After their ``Page \& Screen Unfiltered'' podcast episode receives unexpectedly positive reviews, Ruthanna captures celebratory reactions from ``Bookworms Uncensored'' group chat--including one member stating the 52-minute runtime felt ``just right.'' The screenshot is stored in their private Dropbox as a keepsake and motivation for future episodes. \\ 
\midrule

\textbf{Record Type} & WhatsApp group chat message screenshot from ``Bookworms Uncensored'' shared to a private Dropbox folder \\
\midrule

\textbf{Format} & 
**Highlighted Quotes with Usernames**

Key celebratory or insightful messages are pulled out and attributed to specific group members, interspersed with brief context notes. Tone is appreciative and slightly formal, focusing on memorable remarks.\\ 
\midrule
\textbf{Grouped Attributes} & 
\begin{itemize}[leftmargin=1em,itemsep=-0.3em]
\item \textbf{Ruthanna van der Merwe Personal Identifiers and Contact Information}
\begin{itemize}[itemsep=-0.3em]
\item Ruthanna\_vdm: location: 45 Oxford Rd, Rosebank
\item user\_handle: Ruthanna\_vdm
\item full\_name: Ruthanna van der Merwe
\item whatsapp\_username: Ruthanna\_vdm
\end{itemize}
\item \textbf{Podcast Participants and Collaborators (User Handles and Names)}
\begin{itemize}[itemsep=-0.3em]
\item Ruthanna\_vdm: 
\begin{itemize}[itemsep=-0.3em]
\item location: 45 Oxford Rd, Rosebank
\item user\_handle: Ruthanna\_vdm
\end{itemize}
\item Lindiwe\_J: 
\begin{itemize}[itemsep=-0.3em]
\item user\_handle: Lindiwe\_J
\end{itemize}
\item MphoReads: 
\begin{itemize}[itemsep=-0.3em]
\item user\_handle: MphoReads
\end{itemize}
\item ThaboLitLover:
\begin{itemize}[itemsep=-0.3em]
\item user\_handle: ThaboLitLover
\end{itemize}
\item NalediBooks:
\begin{itemize}[itemsep=-0.3em]
\item user\_handle: NalediBooks
\end{itemize}
\item Sipho: 
\begin{itemize}[itemsep=-0.3em]
\item first\_name: Sipho
\end{itemize}
\item collaborator: Lindiwe
\item sound\_engineer: Sipho
\end{itemize}
\item \textbf{Podcast Episode Metadata (Title, Number, Runtime, Dates)}
\begin{itemize}[itemsep=-0.3em]
\item podcast\_name: Page \& Screen Unfiltered
\item podcast\_episode\_number: 12
\item podcast\_episode\_title: Unreliable Minds
\item podcast\_episode\_runtime: 52-minute
\item podcast\_recording\_date: December 22, 2013
\item podcast\_release\_date: 2014-01-08
\item podcast\_release\_time: 20:17 SAST
\item podcast\_schedule: biweekly Thursday, 19:00 SAST
\end{itemize}
\item \textbf{Podcast Location Information (Recording, Home Studio, Event)}
\begin{itemize}[itemsep=-0.3em]
\item Ruthanna\_vdm:
\begin{itemize}[itemsep=-0.3em]
\item location: 45 Oxford Rd, Rosebank
\item user\_handle: Ruthanna\_vdm
\end{itemize}
\item podcast\_recording\_location: 45 Oxford Rd, Rosebank
\item podcast\_home\_studio: 45 Oxford Rd, Rosebank
\item event\_location: Johannesburg, South Africa
\end{itemize}
\item \textbf{Media and Documentation (Screenshots and Files)}
\begin{itemize}[itemsep=-0.3em]
\item screenshot\_file\_path: \url{/Private/BookwormsUncensored/PodcastFeedback/2014-01-08\_Ep12\_UnreliableMinds\_52min.png}
\end{itemize}
\item \textbf{Podcast Production Roles (Collaborator, Sound Engineer)}
\begin{itemize}[itemsep=-0.3em]
\item collaborator: Lindiwe
\item sound\_engineer: Sipho
\item Sipho: 
\begin{itemize}[itemsep=-0.3em]
\item first\_name: Sipho
\end{itemize}
\end{itemize}
\end{itemize}\\
\bottomrule
\end{tabularx}
\end{table}

\end{document}